\documentclass[runningheads]{llncs}
\usepackage{graphicx}
\usepackage{times}

\makeatletter
\newcommand{\@chapapp}{\relax}%
\makeatother
\usepackage{appendix}

\usepackage{comment}
\usepackage{amsmath,amssymb} %
\usepackage{color}

\usepackage{graphicx}
\usepackage{stmaryrd}
\usepackage{booktabs}
\usepackage{soul}
\usepackage{flushend}
\usepackage{array,multirow}
\usepackage{algpseudocode}
\usepackage{algorithm2e}
\usepackage{comment}
\usepackage{booktabs,tabularx,ragged2e}
\usepackage{multirow}
\usepackage{fontawesome}
\usepackage[table,dvipsnames]{xcolor}

\usepackage{enumitem}
\usepackage{tabu}
\usepackage{ifthen}
\usepackage{siunitx}
\usepackage{pgfplots}

\usepackage[caption=false,position=top]{subfig}
\usepackage{pgfplots}
\pgfplotsset{compat=1.12}
\usepackage{capt-of}

\usepackage[pagebackref=true,breaklinks=true,colorlinks,citecolor=blue,bookmarks=false]{hyperref}

\usepackage[capitalize]{cleveref}
\crefname{section}{Sec.}{Secs.}
\Crefname{section}{Section}{Sections}
\Crefname{table}{Table}{Tables}
\crefname{table}{Tab.}{Tabs.}

\newif\ifnuscenes
\nuscenesfalse

\newcommand{\ours}{Drive$\&$Segment\xspace}
\newcommand{\teacher}{\textsc{T}} %
\newcommand{\student}{\textsc{S}} %
\newcommand{\das}{D\&S\xspace}
\newcommand{\supp}{appendix\xspace}

\newcommand{\paragraphcustom}[1]{\smallskip\noindent\textbf{#1}}

\def\Plus{\texttt{+}}
\def\Minus{\texttt{-}}

\newcolumntype{?}{!{\vrule width 1pt}}

\newcolumntype{d}{>{\tiny}l}

\input{inputs/plots}

\definecolor{customred}{rgb}{0.839, 0.153, 0.157}
\definecolor{LightCyan}{rgb}{0.88,1,1}
\definecolor{bestbg}{gray}{0.90}
\definecolor{aliceblue}{rgb}{0.94, 0.97, 1.0}

\usepackage{hhline}
\usepackage{pgf}
\usepackage{collcell}

\newcommand*{\MinNumber}{0.0}%
\newcommand*{\MidNumber}{0.3} %
\newcommand*{\MaxNumber}{1.0}%

\def\stripzero#1{\expandafter\stripzerohelp#1}
\def\stripzerohelp#1{\ifx 0#1\expandafter\stripzerohelp\else#1\fi}
\newcommand{\ApplyGradient}[1]{%
        \ifdim #1 pt < 0 pt
            \pgfmathsetmacro{\Value}{(-#1)}
            \ifdim \Value pt > \MidNumber pt
                \pgfmathsetmacro{\PercentColor}{max(min(100.0*(\Value - \MinNumber)/(\MaxNumber-\MinNumber),100.0),0.00)} %
                \hspace{-0.96em}
                \setlength{\fboxsep}{-.2pt}
                \colorbox{Goldenrod!\PercentColor!BlueGreen}{\makebox(16,16){\centering \footnotesize \stripzero{\Value}}}  %
            \else
                \pgfmathsetmacro{\PercentColor}{max(min(100.0*(\MidNumber - \Value)/(\MidNumber-\MinNumber),100.0),0.00)} 
                \hspace{-0.96em}
                \setlength{\fboxsep}{-.2pt}
                \colorbox{Violet!\PercentColor!BlueGreen}{\makebox(16,16){\centering \footnotesize \stripzero{\Value}}}%
            \fi    
        \else
            \ifdim #1 pt > \MidNumber pt
                \pgfmathsetmacro{\PercentColor}{max(min(100.0*(#1 - \MinNumber)/(\MaxNumber-\MinNumber),100.0),0.00)} %
                \hspace{-0.65em}
                \setlength{\fboxsep}{-.2pt}
                \colorbox{Goldenrod!\PercentColor!BlueGreen}{\makebox(16,16){\centering \footnotesize \stripzero{#1}}}%
            
            \else
                \pgfmathsetmacro{\PercentColor}{max(min(100.0*(\MidNumber - #1)/(\MidNumber-\MinNumber),100.0),0.00)} %
                \hspace{-0.65em}
                \setlength{\fboxsep}{-.2pt}
                
                \colorbox{Violet!\PercentColor!BlueGreen}{\makebox(16,16){\centering \footnotesize \textcolor{Violet!\PercentColor!BlueGreen}{\stripzero{#1}}}}%
            \fi
        \fi
}
\newcolumntype{G}{>{\collectcell\ApplyGradient}c<{\endcollectcell}}
\usepackage{adjustbox}
\usepackage{array}

\newcolumntype{R}[2]{%
    >{\adjustbox{angle=#1,lap=\width-(#2)}\bgroup}%
    l%
    <{\egroup}%
}
\newcommand*\rotz{\multicolumn{1}{R{0}{-.5em}}}%

\pgfdeclarelayer{bg}
\pgfdeclarelayer{bbg}
\pgfsetlayers{bbg,bg,main}

\makeatletter
\pgfkeys{%
  /tikz/on layer/.code={
    \pgfonlayer{#1}\begingroup
    \aftergroup\endpgfonlayer
    \aftergroup\endgroup
  },
  /tikz/node on layer/.code={
    \gdef\node@@on@layer{%
      \setbox\tikz@tempbox=\hbox\bgroup\pgfonlayer{#1}\unhbox\tikz@tempbox\endpgfonlayer\egroup}
    \aftergroup\node@on@layer
  },
  /tikz/end node on layer/.code={
    \endpgfonlayer\endgroup\endgroup
  }
}
\def\node@on@layer{\aftergroup\node@@on@layer}
\makeatother

\usepackage[accsupp]{axessibility}  %

\begin{document}

\newcommand{\head}[1]{{\smallskip\noindent\textbf{#1}}}
\newcommand{\alert}[1]{{\color{red}{#1}}}
\newcommand{\eq}[1]{(\ref{eq:#1})}
\newcommand{\Th}[1]{\tableheadline{#1}}

\newcommand{\red}[1]{{\color{red}{#1}}}
\newcommand{\blue}[1]{{\color{blue}{#1}}}
\newcommand{\green}[1]{{\color{green}{#1}}}

\definecolor{better}{rgb}{0.19, 0.55, 0.91}
\definecolor{niceblue}{rgb}{0.19, 0.55, 0.91}

\newcommand{\citeme}[1]{\red{[XX]}}
\newcommand{\refme}[1]{\red{(XX)}}

\newcommand{\fig}[2][1]{\includegraphics[width=#1\columnwidth]{\pwd/fig/#2}}
\newcommand{\figh}[2][1]{\includegraphics[height=#1\columnwidth]{\pwd/fig/#2}}

\newcommand{\tran}{^\top}
\newcommand{\mtran}{^{-\top}}
\newcommand{\zcol}{\mathbf{0}}
\newcommand{\zrow}{\zcol\tran}

\newcommand{\ind}{\mathbbm{1}}
\newcommand{\expect}{\mathbb{E}}
\newcommand{\nat}{\mathbb{N}}
\newcommand{\zahl}{\mathbb{Z}}
\newcommand{\real}{\mathbb{R}}
\newcommand{\proj}{\mathbb{P}}
\newcommand{\prob}{\mathbf{Pr}}
\newcommand{\normal}{\mathcal{N}}

\newcommand{\mif}{\textrm{if}\ }
\newcommand{\other}{\textrm{otherwise}}
\newcommand{\minimize}{\textrm{minimize}\ }
\newcommand{\maximize}{\textrm{maximize}\ }

\newcommand{\id}{\operatorname{id}}
\newcommand{\const}{\operatorname{const}}
\newcommand{\sgn}{\operatorname{sgn}}
\newcommand{\var}{\operatorname{Var}}
\newcommand{\mean}{\operatorname{mean}}
\newcommand{\trace}{\operatorname{tr}}
\newcommand{\diag}{\operatorname{diag}}
\newcommand{\vect}{\operatorname{vec}}
\newcommand{\cov}{\operatorname{cov}}
\newcommand{\sign}{\operatorname{sign}}
\newcommand{\prj}{\operatorname{proj}}

\newcommand{\softmax}{\operatorname{softmax}}
\newcommand{\clip}{\operatorname{clip}}

\newcommand{\defn}{\mathrel{:=}}
\newcommand{\peq}{\mathrel{+\!=}}
\newcommand{\meq}{\mathrel{-\!=}}

\newcommand{\floor}[1]{\left\lfloor{#1}\right\rfloor}
\newcommand{\ceil}[1]{\left\lceil{#1}\right\rceil}
\newcommand{\inner}[1]{\left\langle{#1}\right\rangle}
\newcommand{\norm}[1]{\left\|{#1}\right\|}
\newcommand{\abs}[1]{\left|{#1}\right|}
\newcommand{\frob}[1]{\norm{#1}_F}
\newcommand{\card}[1]{\left|{#1}\right|\xspace}
\newcommand{\diff}{\mathrm{d}}
\newcommand{\der}[3][]{\frac{d^{#1}#2}{d#3^{#1}}}
\newcommand{\pder}[3][]{\frac{\partial^{#1}{#2}}{\partial{#3^{#1}}}}
\newcommand{\ipder}[3][]{\partial^{#1}{#2}/\partial{#3^{#1}}}
\newcommand{\dder}[3]{\frac{\partial^2{#1}}{\partial{#2}\partial{#3}}}

\newcommand{\wb}[1]{\overline{#1}}
\newcommand{\wt}[1]{\widetilde{#1}}

\def\xssp{\hspace{1pt}}
\def\ssp{\hspace{3pt}}
\def\msp{\hspace{5pt}}
\def\lsp{\hspace{12pt}}

\newcommand{\cA}{\mathcal{A}}
\newcommand{\cB}{\mathcal{B}}
\newcommand{\cC}{\mathcal{C}}
\newcommand{\cD}{\mathcal{D}}
\newcommand{\cE}{\mathcal{E}}
\newcommand{\cF}{\mathcal{F}}
\newcommand{\cG}{\mathcal{G}}
\newcommand{\cH}{\mathcal{H}}
\newcommand{\cI}{\mathcal{I}}
\newcommand{\cJ}{\mathcal{J}}
\newcommand{\cK}{\mathcal{K}}
\newcommand{\cL}{\mathcal{L}}
\newcommand{\cM}{\mathcal{M}}
\newcommand{\cN}{\mathcal{N}}
\newcommand{\cO}{\mathcal{O}}
\newcommand{\cP}{\mathcal{P}}
\newcommand{\cQ}{\mathcal{Q}}
\newcommand{\cR}{\mathcal{R}}
\newcommand{\cS}{\mathcal{S}}
\newcommand{\cT}{\mathcal{T}}
\newcommand{\cU}{\mathcal{U}}
\newcommand{\cV}{\mathcal{V}}
\newcommand{\cW}{\mathcal{W}}
\newcommand{\cX}{\mathcal{X}}
\newcommand{\cY}{\mathcal{Y}}
\newcommand{\cZ}{\mathcal{Z}}

\newcommand{\vA}{\mathbf{A}}
\newcommand{\vB}{\mathbf{B}}
\newcommand{\vC}{\mathbf{C}}
\newcommand{\vD}{\mathbf{D}}
\newcommand{\vE}{\mathbf{E}}
\newcommand{\vF}{\mathbf{F}}
\newcommand{\vG}{\mathbf{G}}
\newcommand{\vH}{\mathbf{H}}
\newcommand{\vI}{\mathbf{I}}
\newcommand{\vJ}{\mathbf{J}}
\newcommand{\vK}{\mathbf{K}}
\newcommand{\vL}{\mathbf{L}}
\newcommand{\vM}{\mathbf{M}}
\newcommand{\vN}{\mathbf{N}}
\newcommand{\vO}{\mathbf{O}}
\newcommand{\vP}{\mathbf{P}}
\newcommand{\vQ}{\mathbf{Q}}
\newcommand{\vR}{\mathbf{R}}
\newcommand{\vS}{\mathbf{S}}
\newcommand{\vT}{\mathbf{T}}
\newcommand{\vU}{\mathbf{U}}
\newcommand{\vV}{\mathbf{V}}
\newcommand{\vW}{\mathbf{W}}
\newcommand{\vX}{\mathbf{X}}
\newcommand{\vY}{\mathbf{Y}}
\newcommand{\vZ}{\mathbf{Z}}

\newcommand{\va}{\mathbf{a}}
\newcommand{\vb}{\mathbf{b}}
\newcommand{\vc}{\mathbf{c}}
\newcommand{\vd}{\mathbf{d}}
\newcommand{\ve}{\mathbf{e}}
\newcommand{\vf}{\mathbf{f}}
\newcommand{\vg}{\mathbf{g}}
\newcommand{\vh}{\mathbf{h}}
\newcommand{\vi}{\mathbf{i}}
\newcommand{\vj}{\mathbf{j}}
\newcommand{\vk}{\mathbf{k}}
\newcommand{\vl}{\mathbf{l}}
\newcommand{\vm}{\mathbf{m}}
\newcommand{\vn}{\mathbf{n}}
\newcommand{\vo}{\mathbf{o}}
\newcommand{\vp}{\mathbf{p}}
\newcommand{\vq}{\mathbf{q}}
\newcommand{\vr}{\mathbf{r}}
\newcommand{\Vs}{\mathbf{s}}
\newcommand{\vt}{\mathbf{t}}
\newcommand{\vu}{\mathbf{u}}
\newcommand{\vv}{\mathbf{v}}
\newcommand{\vw}{\mathbf{w}}
\newcommand{\vx}{\mathcalmathcal{x}}
\newcommand{\vy}{\mathbf{y}}
\newcommand{\vz}{\mathbf{z}}

\newcommand{\vone}{\mathbf{1}}
\newcommand{\vzero}{\mathbf{0}}

\newcommand{\valpha}{{\boldsymbol{\alpha}}}
\newcommand{\vbeta}{{\boldsymbol{\beta}}}
\newcommand{\vgamma}{{\boldsymbol{\gamma}}}
\newcommand{\vdelta}{{\boldsymbol{\delta}}}
\newcommand{\vepsilon}{{\boldsymbol{\epsilon}}}
\newcommand{\vzeta}{{\boldsymbol{\zeta}}}
\newcommand{\veta}{{\boldsymbol{\eta}}}
\newcommand{\vtheta}{{\boldsymbol{\theta}}}
\newcommand{\viota}{{\boldsymbol{\iota}}}
\newcommand{\vkappa}{{\boldsymbol{\kappa}}}
\newcommand{\vlambda}{{\boldsymbol{\lambda}}}
\newcommand{\vmu}{{\boldsymbol{\mu}}}
\newcommand{\vnu}{{\boldsymbol{\nu}}}
\newcommand{\vxi}{{\boldsymbol{\xi}}}
\newcommand{\vomikron}{{\boldsymbol{\omikron}}}
\newcommand{\vpi}{{\boldsymbol{\pi}}}
\newcommand{\vrho}{{\boldsymbol{\rho}}}
\newcommand{\vsigma}{{\boldsymbol{\sigma}}}
\newcommand{\vtau}{{\boldsymbol{\tau}}}
\newcommand{\vupsilon}{{\boldsymbol{\upsilon}}}
\newcommand{\vphi}{{\boldsymbol{\phi}}}
\newcommand{\vchi}{{\boldsymbol{\chi}}}
\newcommand{\vpsi}{{\boldsymbol{\psi}}}
\newcommand{\vomega}{{\boldsymbol{\omega}}}

\newcommand{\rLambda}{\mathrm{\Lambda}}
\newcommand{\rSigma}{\mathrm{\Sigma}}

\newcommand{\vLambda}{\bm{\rLambda}}
\newcommand{\vSigma}{\bm{\rSigma}}

\makeatletter
\newcommand*\bdot{\mathpalette\bdot@{.7}}
\newcommand*\bdot@[2]{\mathbin{\vcenter{\hbox{\scalebox{#2}{$\m@th#1\bullet$}}}}}
\makeatother

\makeatletter
\DeclareRobustCommand\onedot{\futurelet\@let@token\@onedot}
\def\@onedot{\ifx\@let@token.\else.\null\fi\xspace}
\def\eg{\emph{e.g}\onedot} \def\Eg{\emph{E.g}\onedot}
\def\ie{\emph{i.e}\onedot} \def\Ie{\emph{I.e}\onedot}
\def\cf{\emph{cf}\onedot} \def\Cf{\emph{C.f}\onedot}
\def\etc{\emph{etc}\onedot} \def\vs{\emph{vs}\onedot}
\def\wrt{w.r.t\onedot} \def\dof{d.o.f\onedot} \def\aka{a.k.a\onedot}
\def\etal{\emph{et al}\onedot}
\makeatother

\usetikzlibrary{arrows.meta, shadows, fadings,shapes.arrows}
\tikzset{Arrow Style/.style={single arrow, fill=orange, anchor=base, align=center,text width=.3cm}}
\newcommand{\MyArrow}[2][]{\tikz[baseline] \node [My Arrow Style,#1] {#2};}

\pagestyle{headings}
\mainmatter

\title{Drive\&Segment: Unsupervised Semantic Segmentation of Urban Scenes via Cross-modal Distillation} 
\titlerunning{\ours}
\author{Antonin Vobecky\inst{1,2} \and
David Hurych\inst{2}\and
Oriane Sim\'eoni\inst{2}\and
Spyros Gidaris\inst{2}\and
Andrei Bursuc\inst{2}\and
Patrick P\'erez\inst{2}\and
Josef Sivic\inst{1}
}
\authorrunning{Vobecky et al.}
\institute{Czech Institute of Informatics, Robotics and Cybernetics, CTU in Prague \and
valeo.ai}

\colorlet{bestbg}{niceblue!20!white}

\maketitle

\begin{abstract}
This work investigates learning pixel-wise semantic image segmentation in urban scenes without any manual annotation, just from the raw non-curated data collected by cars which, equipped with cameras and LiDAR sensors, drive around a city. Our contributions are threefold. First, we propose a novel method for cross-modal unsupervised learning of semantic image segmentation by leveraging synchronized LiDAR and image data. The key ingredient of our method is the use of an object proposal module that analyzes the LiDAR point cloud to obtain proposals for spatially consistent objects. Second, we show that these 3D object proposals can be aligned with the input images and reliably clustered into semantically meaningful pseudo-classes. Finally, we develop a cross-modal distillation approach that leverages image data partially annotated with the resulting pseudo-classes to train a transformer-based model for image semantic segmentation. 
We show the generalization capabilities of our method by testing on four different testing datasets (Cityscapes, Dark Zurich, Nighttime Driving and ACDC) without any finetuning, and demonstrate significant improvements compared to the current state of the art on this problem.
\footnote{See project webpage \href{https://vobecant.github.io/DriveAndSegment/}{https://vobecant.github.io/DriveAndSegment/} for the code and more.}

\keywords{autonomous driving \and unsupervised semantic segmentation}
\end{abstract}

\section{Introduction}
\label{sec:intro}

In this work, we investigate whether it is possible to learn pixel-wise semantic image segmentation of urban scenes without the need for any manual annotation, just from the raw non-curated data that are collected by cars equipped with cameras and LiDAR sensors while driving in town. 
This topic is important as current methods require large amounts of pixel-wise annotations over various driving conditions and situations. Such a manual segmentation of images on a large scale is very expensive, time consuming, and prone to biases.

\begin{figure}[t]
    \centering
    \includegraphics[width=1.0\linewidth]{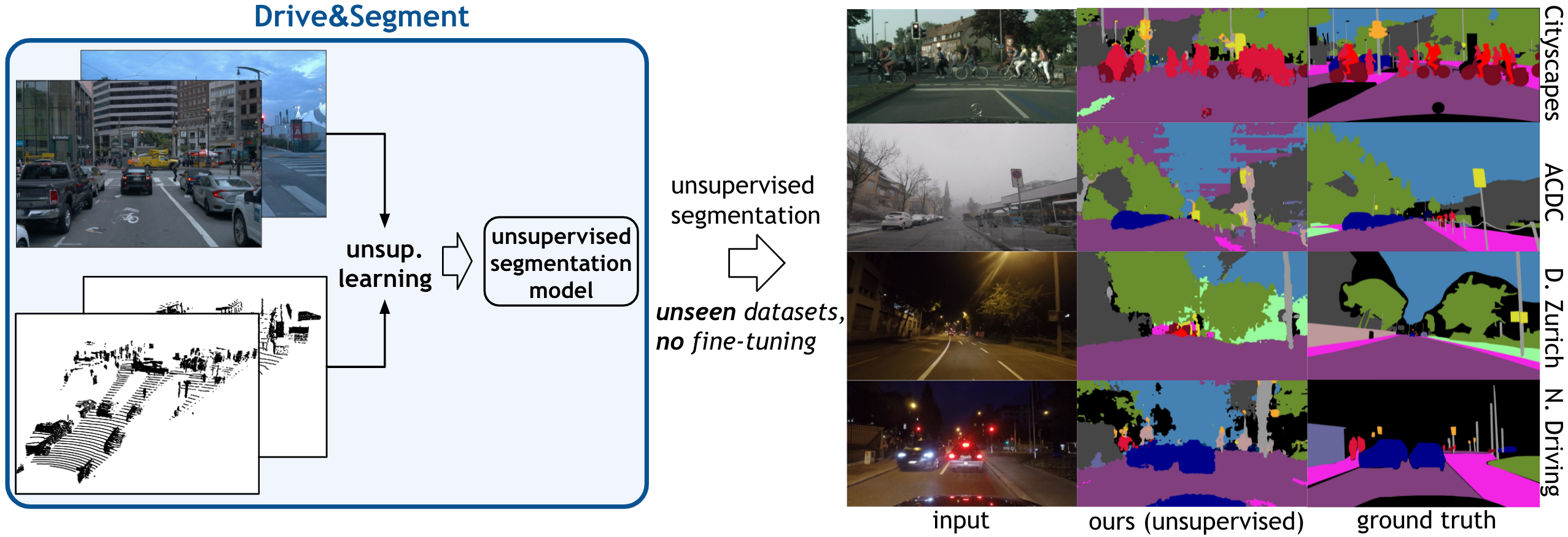}
    \vspace{-20pt}
    \caption{
    {\bf Proposed fully-unsupervised approach.} From uncurated images and LiDAR data, our \ours approach learns a semantic image segmentation model with no manual annotations. The resulting model performs unsupervised semantic segmentation of new unseen datasets 
    without any human labeling. It can segment complex scenes with many objects, including thin structures such as \textcolor{red}{people}, \textcolor{purple}{bicycles}, \textcolor{gray}{poles} or \textcolor{orange}{traffic lights}. Black denotes the ignored label.
    }
    \label{fig:teaser}
    \vspace{-15pt}
\end{figure}

Currently, the best methods for unsupervised learning of semantic segmentation assume that images contain centered objects~\cite{van2021unsupervised} rather than full scenes, or use spatial self-supervision available in the image domain~\cite{cho2021picie}. They do not leverage additional modalities, such as the LiDAR data, available for urban scenes in the autonomous driving set-ups. In this work, we develop an approach for unsupervised semantic segmentation that learns to segment complex scenes containing many objects, including thin structures such as pedestrians or traffic lights, without the need for any manual annotation, but instead leveraging cross-modal information available in (aligned) LiDAR point clouds and images, see Fig.~\ref{fig:teaser}. 
Exploiting point clouds as a form of supervision is, however, not straightforward: data from LiDAR and camera are rarely perfectly synchronized; moreover, point clouds are unstructured and of much lower resolution compared to RGB images; finally, extracting semantic information from LiDAR is still a very hard problem. In this work, we overcome these issues and demonstrate that it is nevertheless possible to extract useful pixel-wise semantic supervision from LiDAR data.

The contributions of our work are threefold. First, we propose a novel method for cross-modal unsupervised learning of semantic image segmentation by leveraging synchronized LiDAR and image data. The key ingredient is a module analyzing the LiDAR point cloud to obtain proposals for spatially consistent objects that can be clearly separated from each other and from the ground plane in the 3D scene. 
Second, we show that these 3D object proposals can be aligned with input images and reliably clustered into semantically meaningful pseudo-classes 
by using image features from a network trained without supervision.
We demonstrate that this approach is robust to noise in point clouds and delivers, without the need for any manual annotation, pseudo-classes with pixel-wise segmentation for a variety of objects present in driving scenes. These classes include objects such as pedestrians or traffic lights that are notoriously hard to segment automatically in the image domain. Third, we develop a novel cross-modal distillation approach that first trains a teacher network with the available partial pseudo labels, and then exploits its predictions for training the student with pixel-wise pseudo annotations 
that cover the whole image.
In addition, our approach exploits geometric constraints extracted from the LiDAR point cloud during the teacher-student learning process to refine the teacher predictions that are distilled into the student network. Implemented with transformer-based networks, this cross-modal distillation approach 
results in a trained student model that performs well in a variety of challenging conditions such as day, night, fog, or rain, outside the domain of the original training dataset, as shown in Fig.\,\ref{fig:teaser}.

We train our proposed  unsupervised  semantic  segmentation  method  on Waymo Open~\cite{sun2020scalability} and nuScenes~\cite{nuscenes} datasets (nuScenes results are in the 
\supp
), and test it on four different datasets in the autonomous driving domain, Cityscapes~\cite{Cordts2016Cityscapes}, DarkZurich~\cite{SDV20}, Nighttime driving~\cite{daytime:2:nighttime} and ACDC~\cite{SDV21}. We demonstrate significant improvements compared to the current state of the art, improving the current best published unsupervised semantic segmentation results on Cityscapes from $15.8$ to $21.8$ and from $4.6$ to $14.2$ on Dark Zurich, measured by mean intersection over union.

\begin{figure*}[t!]
    \centering
    \resizebox{\linewidth}{!}{\input{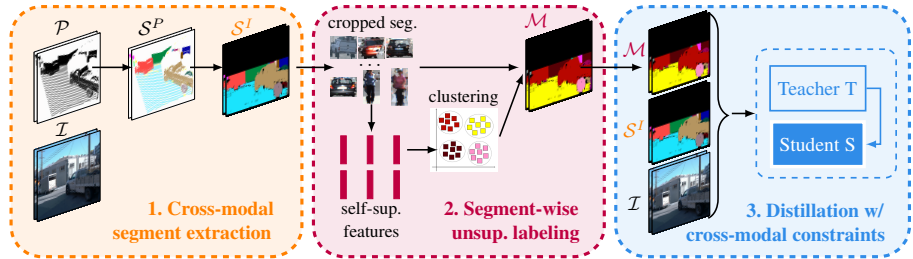}}
    \vspace*{-18pt}
    \caption{
    {\bf Overview of %
    \ours.  
    }
    We first perform 
    \textcolor{orange}{cross-modal segment extraction}
    on training dataset by exploiting raw \emph{LiDAR} point clouds $\cP$ and raw \emph{images} $\cI$. This yields segments $\cS^I$ projected onto the image space (\S\,\ref{sec:segment_extraction}). By clustering their self-supervised features, we obtain an  \textcolor{purple}{unsupervised labeling of these segments} (\S\,\ref{sec:clustering}) and, as a consequence, of their pixels. This provides pixel-wise \emph{pseudo ground truth} for the next learning step.  
    Finally, given the pseudo-labels and the segments, we perform \textcolor{niceblue}{distillation with cross-modal constraints} (\S\,\ref{sec:distillation}) that 
    conjugates information of the LiDAR and the images to learn a final segmentation model using a teacher-student architecture. The learnt segmentation model \student{} --highlighted in the figure-- is used for inference on unseen datasets, yielding compelling results (\S\,\ref{sec:experiments}).
    }
    \label{fig:pipeline}
    \vspace*{-3mm}
\end{figure*}

\section{Related work}

In this work, we investigate the task of \emph{semantic image segmentation with no human supervision}. We discuss the corresponding prior art below.

\paragraphcustom{Image semantic segmentation.}
Semantic segmentation is a challenging key visual perception task, especially for autonomous driving~\cite{Cordts2016Cityscapes,neuhold2017mapillary,SDV21,varma2019idd,yu2020bdd100k}. Current top-performing models are based on fully convolutional networks~\cite{long2015fully} with encoder-decoder structures and a large diversity of designs~\cite{chen2018encoder,Cheng_2020_CVPR,lin2017feature,Ronneberger2015UNet,wang2020deep,zhao2017pyramid}. 
Recent progress in vision transformers (ViT)~\cite{dosovitskiy2021image} opened the door for a new wave of decoders~\cite{Strudel_2021_ICCV,xie2021segformer,yuan2020object,Zheng_2021_CVPR} with appealing performance. All methods, in particular transformer-based~\cite{dosovitskiy2021image}, attain impressive performance by exploiting large amounts of pixel-wise labeled data. Yet, %
urban scenes are expensive to annotate manually ($1.5$h-$3$h per image~\cite{Cordts2016Cityscapes,SDV21}). This motivates recent works to rely less on pixel-wise supervision.

\paragraphcustom{Reducing supervision for semantic segmentation.}
A popular strategy when dealing with limited labeled data is to pre-train some of the blocks of the architecture, e.g., the encoder, on related auxiliary tasks with plentiful labels~\cite{deng2009imagenet,zamir2018taskonomy}. Pre-training encoder for image classification, e.g., on ImageNet~\cite{deng2009imagenet}, has been shown to be a successful recipe for both convnets~\cite{chen2018encoder} and ViT-based models~\cite{Strudel_2021_ICCV}. Pre-training can be conducted even without any human annotations on artificially-designed self-supervised pretext tasks~\cite{caron2018deep,gidaris2021obow,Gidaris2018Unsupervised,grill2020bootstrap,he2020momentum,henaff2021efficient} with impressive results on a variety of downstream tasks.
Fully unsupervised semantic segmentation has been recently addressed ~\cite{bielski2019emergence,chen2019unsupervised,cho2021picie,hwang2019segsort,ji2019invariant,kanezaki2018unsupervised,ouali2020autoregressive,van2021unsupervised,zhang2020self} via generative models for generating object masks~\cite{bielski2019emergence,chen2019unsupervised,ouali2020autoregressive} or self-supervised clustering~\cite{cho2021picie,ji2019invariant}. 
Prior methods are limited to segmenting foreground objects of a single class~\cite{bielski2019emergence,chen2019unsupervised} or to \emph{stuff} pixels that outnumber by far \emph{things} pixels~\cite{ji2019invariant,ouali2020autoregressive}. Others assume that images contain centered objects~\cite{van2021unsupervised}, rely on weak spatial cues from the image domain~\cite{chen2019unsupervised,cho2021picie,ji2019invariant} or require instance masks during pre-training and annotated data at test time~\cite{hwang2019segsort}. In contrast, our approach exploits cross-modal supervision from aligned LiDAR point clouds and images. We show that leveraging this information can considerably improve segmentation performance in complex autonomous driving scenes with multiple classes and strong class imbalance, outperforming PiCIE~\cite{cho2021picie}, the current state of the art in unsupervised segmentation.

\paragraphcustom{Cross-modal self-supervised learning.}
Leveraging language, vision, and/or audio,  self-supervised representation learning has seen tremendous progress in recent years~\cite{alayrac2020self,alwassel_2020_xdc,arandjelovic2017look,miech2020end,owens2018audio,Recasens_2021_ICCV,zhao2018sound}. Besides learning useful representations, these approaches show that signals from one modality can help train object detectors in the other, e.g., detecting instruments that sound in a scene~\cite{chen2021localizing,owens2018audio,zhao2018sound}, and even other object types~\cite{afouras2021self}. In autonomous driving, a vehicle is equipped with diverse sensors (e.g., camera, LiDAR, radar) and cross-modal self-supervision is often used to generate labels from a sensor for augmenting the perception of another~\cite{bartoccioni2021lidartouch,jaritz2020xmuda,tian2021unsupervised,weston2019probably}. LiDAR clues~\cite{tian2021unsupervised} have been recently shown to improve unsupervised object detection. In contrast, we consider the problem of pixel-wise unsupervised semantic segmentation, a particularly challenging task given the sparsity and low resolution of LiDAR point clouds.

\section{Proposed unsupervised semantic segmentation}

Our goal is to train an \emph{image segmentation model} with \textit{no human annotation}, by exploiting easily-available aligned \emph{LiDAR} and \emph{image} data. 
To that end, we propose a novel method, \ours, that consists of three major steps and is illustrated in Figure~\ref{fig:pipeline}.
First, as discussed in Section~\ref{sec:segment_extraction}, we extract \emph{segment} proposals for the objects of interest from 3D LiDAR point clouds and project them to the aligned RGB images. In the second step, presented in Section~\ref{sec:clustering}, we build \emph{pseudo-labels} by clustering \emph{self-supervised} image features corresponding to these segments. Finally, in Section~\ref{sec:distillation}, we propose a new teacher-student training scheme that incorporates \emph{spatial constraints} from the LiDAR data and learns an unsupervised segmentation model from the noisy and partial pseudo-annotations generated in the previous two steps.

\subsection{Cross-modal segment extraction}
\label{sec:segment_extraction}

\begin{figure*}[t]
    \centering
    \vspace{-5pt}
    \includegraphics[width=1.0\textwidth]{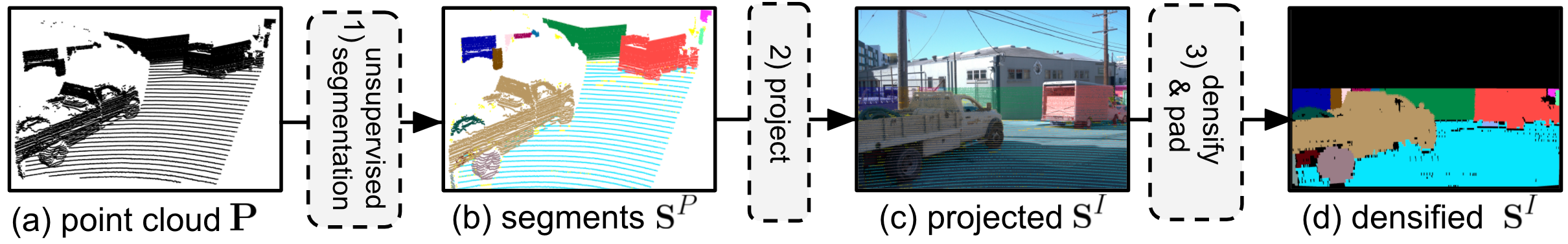}
    \vspace{-22pt}
    \caption{\textbf{Cross-modal segment extraction}. Input raw point cloud (a) is first segmented with~\cite{bogoslavskyi17pfg}
    into object segment candidates (b), which are then projected into the image (c); Projected segments are densified to get pixel-level pseudo-labels, with missed pixels being labeled as ``ignore'', as shown in black (d).
    }
    \label{fig:segment_extraction_row}
    \vspace*{-6mm}
\end{figure*}

Throughout the next sections, we consider a dataset composed of a set $\cP$ of 3D point clouds and a set $\cI$ of images aligned with the point clouds. In this section, we detail the process for extracting segments of interest in an image $\vI \in \cI$ using the corresponding aligned LiDAR point cloud $\vP \in \cP$. The process, illustrated in Fig.\,\ref{fig:segment_extraction_row}, consists of three major steps. We start by segmenting the LiDAR point cloud 
$\vP$ using its geometrical properties. Then, we project the resulting 3D segments into the image
$\vI$, and densify the output to obtain pixel-level segments. 

\paragraphcustom{Geometric point cloud segmentation.} 
We first extract $J$ non-overlapping object segmentation proposals %
(\emph{segments}), from the LiDAR point cloud
$\vP$. Let $\vS^P{=}\{s^P_{j}\}_{j=1}^{J}$ be this set, %
where each segment $s^P_{j}$ %
is a subset of %
the 3D point cloud $\vP$ and, ${\forall j \neq j',\ s^P_{j} \cap s^P_{j'} = \varnothing}$. Additionally, we refer to the set of segments over the entire dataset as $\cS^P$, with $\vS^P \subset \cS^P$. The $J$ segments detected in one point cloud should ideally correspond to $J$ individual objects in the scene. To get them, we use the unsupervised 3D point cloud segmentation proposed in \cite{bogoslavskyi17pfg}, which exploits the geometrical properties of point clouds and range images.
\footnote{
Range images are depth maps corresponding to the raw LiDAR measurements. Valid measurements are back-projected to the 3D space to form a point cloud.
} 
It is a two-stage process that segments the ground plane and objects using greedy labeling by breadth-first search in the range image domain. Urban scenes are particularly suited to this purely geometry-based method as most objects are spatially well separated and the ground plane %
is relatively easy to segment out.

\paragraphcustom{Point-cloud-to-image transfer.} 
The next step of the segment extraction is to transfer the set $\vS^P$ of point cloud segments 
to the image~$\vI$,
producing the set 
$\vS^I$. Despite LiDAR data and camera images being captured at the same time, one-to-one matching is not straightforward. 
Indeed, among other difficulties, LiDAR data only covers a fraction of the image plane because of its different field of view, its lower density and its lack of usable measurements on far away objects or on the sky for instance.
To overcome the mismatch between the two modalities, we proceed as follows.  
First, we project the points from the point cloud to the image using the known sensors' calibration. This gives us the locations of 3D points from the point cloud in the image.  
We also identify locations with invalid measurements in the LiDAR's range image, 
e.g., reflective surfaces or the sky, 
and assign an ``ignore'' label to the respective locations.

\paragraphcustom{Densify \& pad.} 
Next, we perform nearest-neighbor interpolation to propagate the $J+1$ segment labels to all pixels, where $J$ is the number of segments (ideally corresponding to objects) and $+1$ denotes the additional ``ignore'' label. Last, we pad the image with ``ignore'' label to the input image size.

\subsection{Segment-wise unsupervised labeling}
\label{sec:clustering}

\begin{figure}[t]
    \centering
    \includegraphics[width=1.0\textwidth]{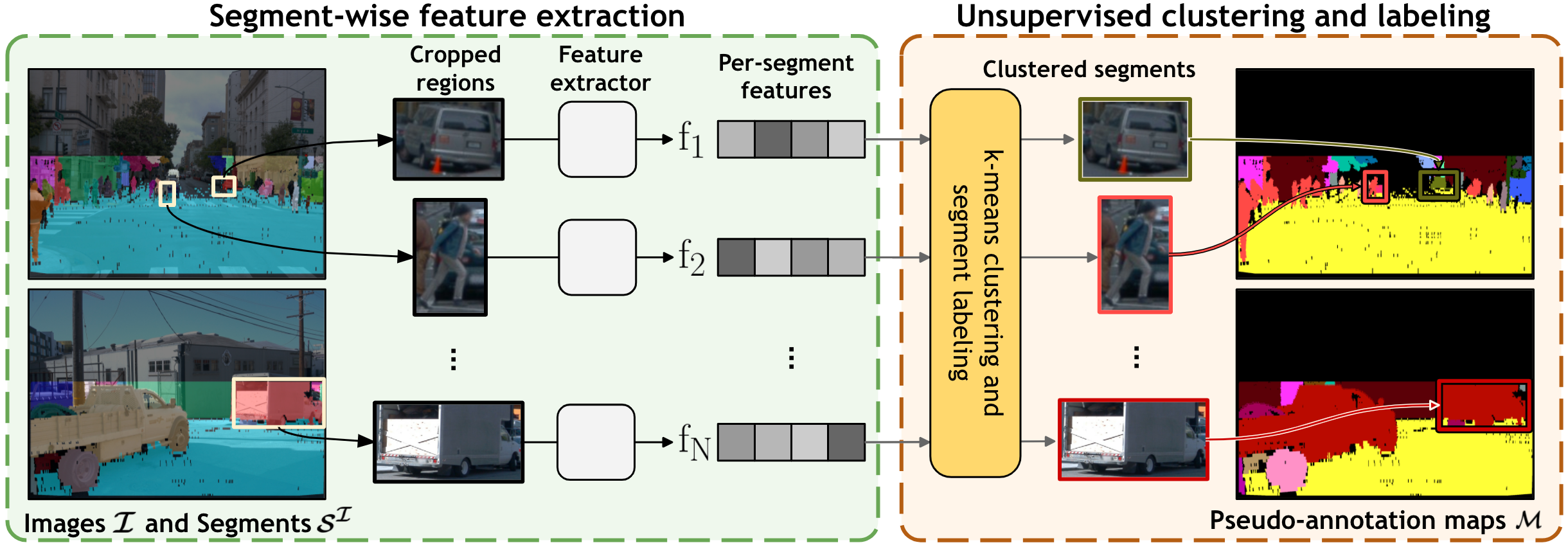}
    \vspace{-5ex}
    \caption{\small 
    \textbf{Segment-wise unsupervised pseudo-labeling}. First, given object segments $\mathcal{S^I}$ obtained in the segment extraction stage (left), we take crops around all $N$ objects and feed them to 
    a
    feature extractor to get a set of $N$ feature vectors. Then, we use the $k$-means algorithm to cluster the feature vectors into $k$ clusters. Finally, we assign pixel-wise {\em pseudo-labels} to all pixels belonging to each segment based on the corresponding cluster id. Pixels not covered by a segment are assigned the label ``ignore" (black).
    }
    
    \label{fig:labeling pipeline}
    \vspace{-3ex}
\end{figure}

Next, the objective is to produce \emph{pseudo-labels} for all extracted segments of interest in the image space, and to do so without any supervision. In particular, we leverage the very recent ViT~\cite{dosovitskiy2021image} model pre-trained in a fully unsupervised fashion~\cite{caron2021emerging} which has shown impressive results on a range of downstream tasks. We use this representation for unsupervised learning of pseudo-labels as described next and illustrated in Figure~\ref{fig:labeling pipeline}. 

Considering here the image $\vI$, we crop a tight rectangular region in the image around each segment $s^I_{j}$ $\in \vS^I$ obtained using the proposal mechanism described in the previous section. We resize it and feed it to the ViT model to extract the feature $\vf_{j}$ corresponding to the output features of the \texttt{CLS} token. To limit the influence of pixels outside the object segment, which may correspond to other objects or the background, we mask out these pixels before computing the features. We repeat this operation for all segments in each image $\vI$ in the training dataset and cluster the \texttt{CLS} token features using $k$-means algorithm, thus discovering $k$ clusters of visually similar segments. Therefore, each feature $\vf_{j}$ and its corresponding segment $s^I_{j}$, is assigned a cluster id $l_{j}$ in $\llbracket 1, k\rrbracket$.

To get a dense \emph{segmentation} map $\vM$ corresponding to the image $\vI$, we assign discovered cluster ids to each pixel belonging to a segment in the image. We additionally assign a predefined \emph{ignore} label to pixels not covered by segments, which correspond to missing annotations. This allows us to construct a set $\cM$ of dense \emph{maps of pseudo-annotations}, that we later use as a pseudo-ground-truth. Examples of resulting segmentation maps are shown in~Figure~\ref{fig:labeling pipeline}.

\subsection{\mbox{Distillation with cross-modal spatial constraints}}
\label{sec:distillation}

After previous steps, we now have a set of pseudo-annotated segmentation maps $\vM \in \cM$, one for every image $\vI$ in the training dataset. However, as explained above, the pseudo-annotations are \textbf{only partial}. This is because the segments that were used to construct them do not cover all pixels of an image. Furthermore, due to imperfections in the segment extraction process or the segment clustering step, these annotations are noisy. Therefore, directly training an image segmentation model with those pseudo-labels might be sub-optimal. Instead, we propose a new teacher-student training approach with cross-modal distillation, which is able to learn more accurate unsupervised segmentation models under such partial and noisy pseudo-annotations. 

\paragraphcustom{Training the teacher.} The first step of our teacher-student approach is to train the teacher~$\teacher $ to make pixel-wise predictions only on the pixels for which pseudo-annotations are available, i.e., only for the pixels that belong to a segment. We denote $\vY_{\teacher } = \teacher \left( \vI \right) \in \real^{H\times W}$ the segmentation predictions made by the teacher model on image $\vI$ with a resolution of $H\times W$ pixels. We train the teacher~$\teacher $ using loss $\mathcal{L}_\teacher (\vI) $ on image $\vI$:  
\begin{equation}
     \mathcal{L}_\teacher (\vI) = \frac{1}{\sum\nolimits_{h,w}{B_{(h,w)}}} 
     \sum_{h,w}
     \mathrm{CE}\left( \vY_{\teacher,(h,w) }, \vM_{(h,w)} \right) %
     B_{(h,w)},
     \label{eq:disti}
\end{equation}
where CE is the cross-entropy loss measuring the discrepancy between the predicted labels $\vY_{\teacher}$ and target pseudo-labels $\vM$ for each pixel $(h,w)$, 
and %
$B$ is a $H\,{\times}\,W$ binary mask for filtering out pixels without pseudo-annotations. %
The loss is normalized w.r.t. the number of pseudo-labeled pixels in the image.
The trained teacher~$\teacher $ is then able to predict pixel-wise segmentation for all pixels in an image, even if they do not belong to a segment. Moreover, since the teacher~$\teacher $ is trained on a large set of pseudo-annotated segments, it learns to smooth out some of the noise in the raw pseudo-annotations.

\begin{figure}[t]
    \centering
    \includegraphics[width=1.\textwidth]{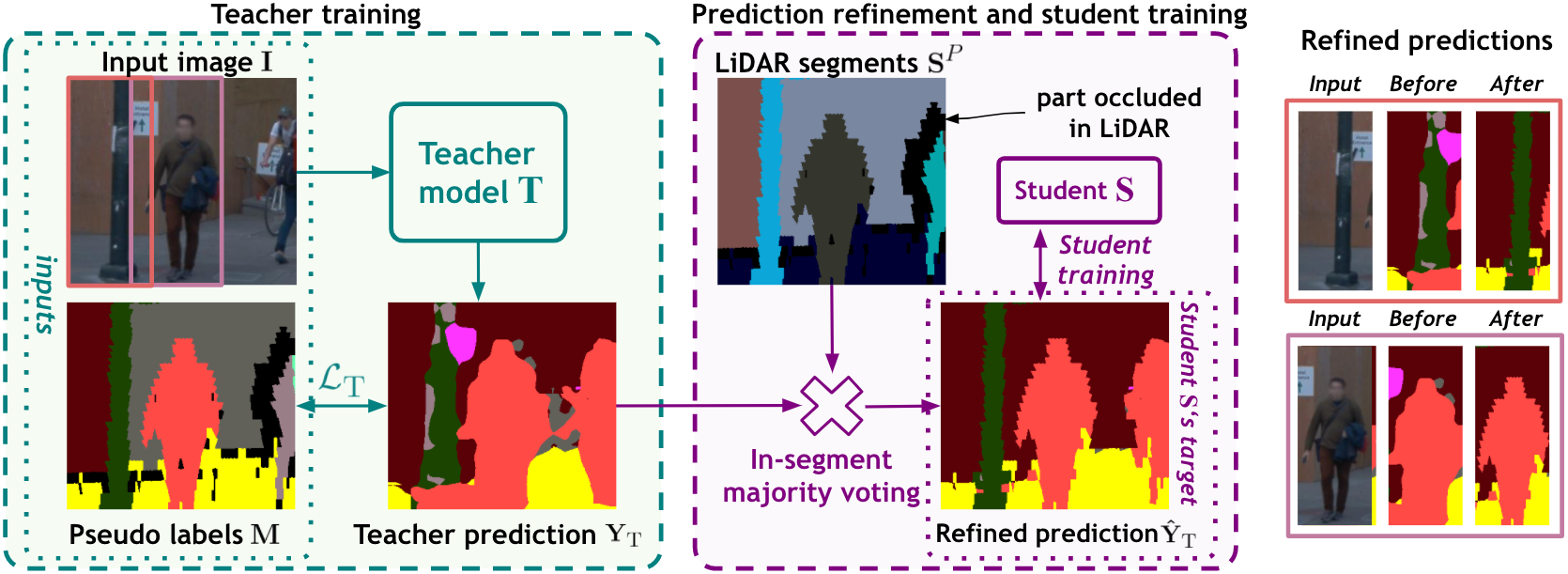}
    \vspace{-20pt}
    \caption{
    \textbf{Teacher prediction refinement using spatial constraints.}
    First, the teacher~$\teacher $ is trained using loss $\mathcal{L}_{\teacher} $ on images in $\cI$ together with segmentation maps in $\cM$ obtained from segment-wise unsupervised pseudo-labeling. The teacher predictions $\mathrm{\vY}_{\teacher }$ are refined, using LiDAR segments $\vS^{P}$, into maps $\mathrm{\hat{\vY}}_{\teacher }$ that are then used to train the student.  
    Note that teacher's  predictions span the whole image, producing outputs even in areas where LiDAR segments $\vS^P$ are not available.
    }
    \label{fig:refinement}
    \vspace{-3ex}
\end{figure}

\paragraphcustom{Integrating spatial constraints.} Considering this smoothing property, we can exploit the trained teacher~$\teacher$ for generating new, complete (instead of partial) and smooth pseudo-segmentation maps for the training images. In addition, we propose to refine these teacher-generated pseudo-segmentation maps by using the projected LiDAR segments; indeed, these segments encode useful 3D spatial constraints as they often correspond to complete 3D objects, thus respecting the depth discontinuities and occlusion boundaries. In particular, for each image segment $s^I_{j}$ in image $\vI$, we apply majority voting to pixel-wise teacher predictions $\mathrm{\vY}_{\teacher }$ inside the segment. Then, we annotate each pixel belonging to the segment with its most frequently predicted label, giving us a new refined segmentation map $\mathrm{\hat{\vY}}_{\teacher } \in \real^{H\times W}$. This procedure is illustrated in Figure~\ref{fig:refinement}.

\paragraphcustom{Training the student.} Having computed these complete, teacher-generated and spatially-refined pseudo-segmentation maps $\mathrm{\hat{\vY}}_{\teacher }$, we  train a 
student network~$\student$ using the following loss
\begin{equation}
     \mathcal{L}_{\text{distill}}(\vI) = \frac{1}{H W} 
     \sum_{h,w}
        \mathrm{CE} \left( 
        \mathrm{\hat{\vY}}_{\teacher,(h,w) }, \mathrm{\vY}_{\student,(h,w) } 
        \right)
    , 
\label{eq:disti_stud}
\end{equation}
where the cross-entropy is computed between 
$\mathrm{\hat{\vY}}_{\teacher }$ and 
the segmentation map~$\mathrm{\vY}_{\student } \in \real^{H\times W}$ predicted by the student at the same resolution as the teacher. %
The outputs of the trained student are our final unsupervised image segmentation predictions. Further details about our training can be found in Section~\ref{sec:implementation_details}.

\section{Experiments}
\label{sec:experiments}
In this section, we give the implementation details, compare our results with the state-of-the-art unsupervised semantic segmentation methods on four different datasets, and ablate the key components of our approach.

\subsection{Experimental setting}
\label{sec:implementation_details}
\begin{figure*}[t]
    \centering
    \begin{minipage}[t]{0.98\columnwidth}
        \footnotesize
         \centering
         \begin{tabular}{c@{}c@{}c}
         Input & Ground Truth & Drive\&Segment (Ours) \\
         \vspace{-3pt}
         \includegraphics[width=0.33\textwidth]{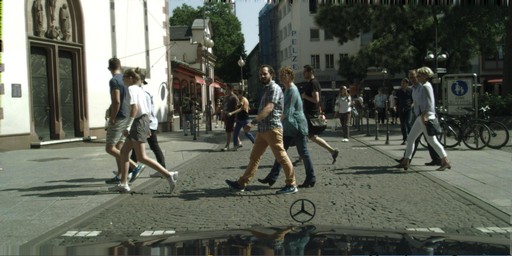} &
         \includegraphics[width=0.33\textwidth]{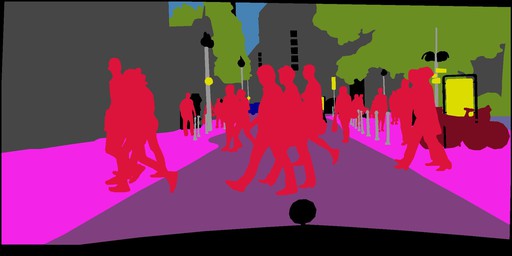} &
         \includegraphics[width=0.33\textwidth]{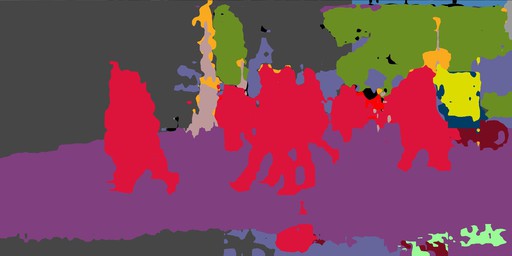} \\
         \vspace{-3pt}
         \includegraphics[width=0.33\textwidth]{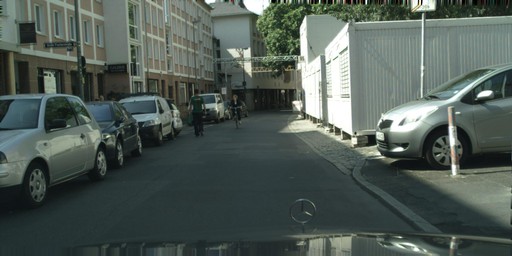} &
         \includegraphics[width=0.33\textwidth]{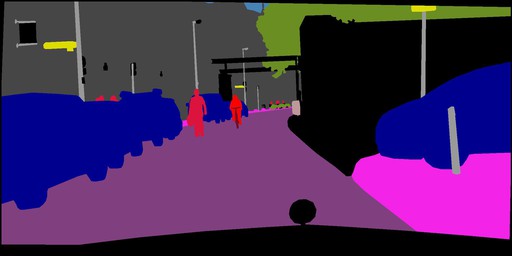} &
         \includegraphics[width=0.33\textwidth]{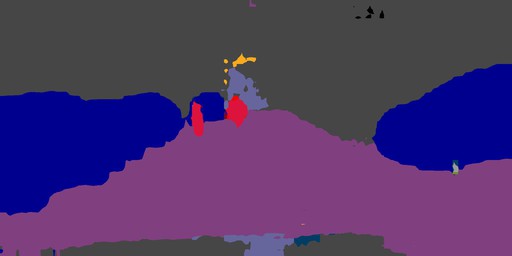} \\
         \vspace{-3pt}
         \includegraphics[width=0.33\textwidth]{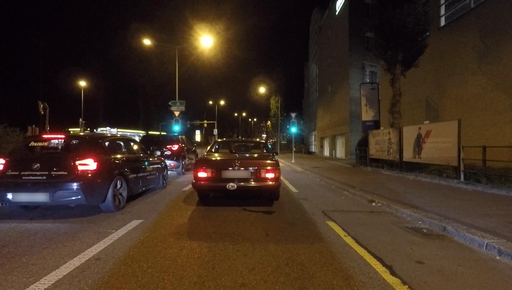} &
         \includegraphics[width=0.33\textwidth]{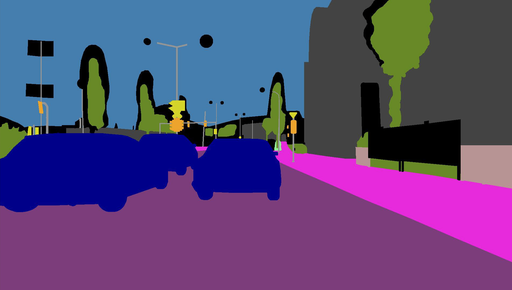} &
         \includegraphics[width=0.33\textwidth]{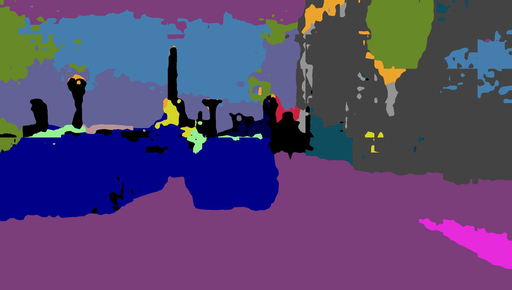} \\
         \vspace{-3pt}
         \includegraphics[width=0.33\textwidth]{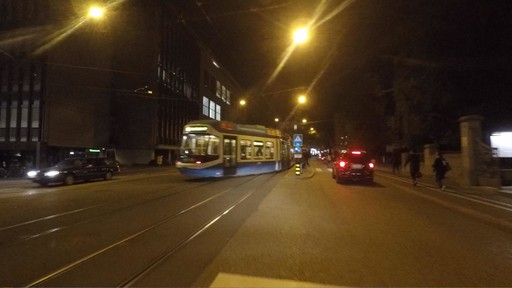} &
         \includegraphics[width=0.33\textwidth]{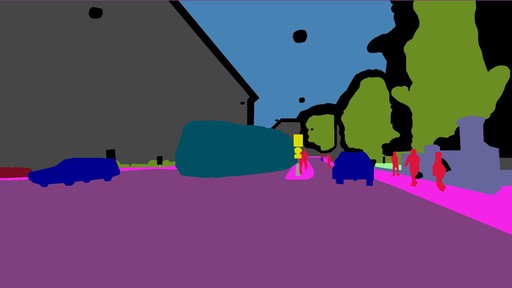} &
         \includegraphics[width=0.33\textwidth]{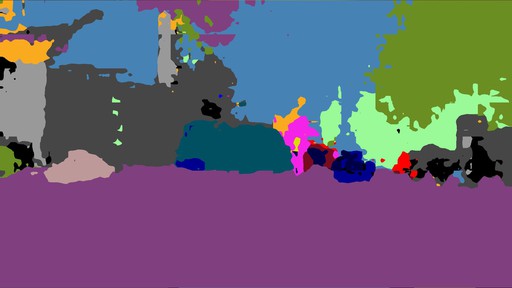} \\
         \vspace{-3pt}
         \includegraphics[width=0.33\textwidth]{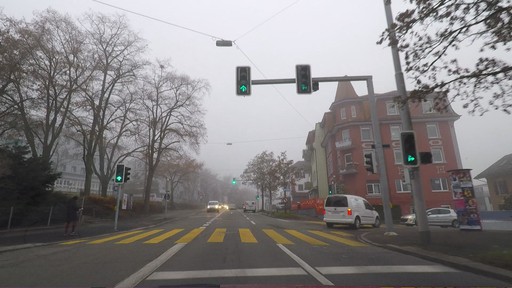} &
         \includegraphics[width=0.33\textwidth]{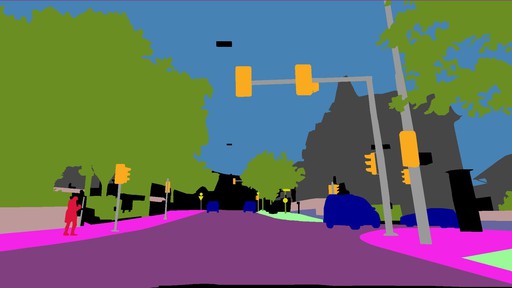} &
         \includegraphics[width=0.33\textwidth]{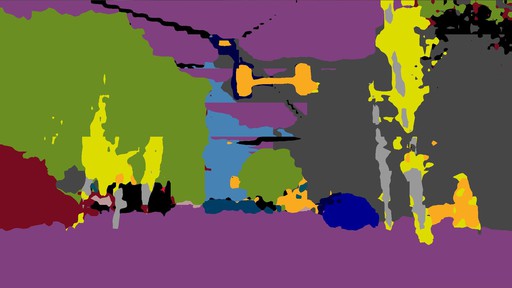} \\
         \end{tabular}
    \end{minipage}\\
    \includegraphics[width=0.98\linewidth]{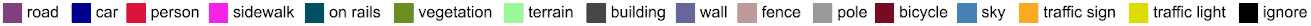}
    \vspace{-6pt}
    \caption{
        \textbf{Qualitative results} for \emph{unsupervised} semantic segmentation using our \ours approach. To obtain the best matching between our pseudo-classes and the set of ground-truth classes, we use the Hungarian algorithm. The first two rows show samples from the Cityscapes~\cite{Cordts2016Cityscapes} dataset, and the other three rows show samples from the night and fog splits of the ACDC~\cite{SDV21} dataset. Please see Figures~\ref{fig:qualitative_supp},~\ref{fig:qualitative_supp2} and~\ref{fig:qualitative_acdc} in the 
        \supp
        for more qualitative results.
    }
    \label{fig:qualitative}
    \vspace{-4ex}
\end{figure*}

\paragraphcustom{Methods and architectures.}
\label{sec:architecture}
We investigate the benefits of our approach using two different semantic segmentation models to demonstrate the generality of our method. We implement \ours with both a classical convolutional model and a transformer-based architecture. For the convolutional architecture, we follow \cite{cho2021picie} and use a ResNet18~\cite{he2016deep} backbone followed by an FPN~\cite{lin2017feature} decoder. For the transformer-based architecture, we use the state-of-the-art Segmenter~\cite{Strudel_2021_ICCV} model.
We use the ViT-S/16~\cite{dosovitskiy2021image} model as the Segmenter's encoder and use a single layer of the mask transformer~\cite{Strudel_2021_ICCV} as a decoder.
We compare our method %
to three recent unsupervised segmentation models: IIC~\cite{ji2019invariant}, modified version of DeepCluster~\cite{caron2018deep} (DC), and PiCIE~\cite{cho2021picie}.
Please refer to \cite{cho2021picie} for implementation details of IIC and DC.

\paragraphcustom{Training.}
\label{sec:training_details}
In the following, we first discuss how we obtain segment labels by k-means clustering, then we talk about details of pre-training the model backbones, which is followed by the discussion of the datasets for actual training of the models. Finally, we give details of the training procedure.

\emph{K-means.} We use $k=30$ in the k-means algorithm (the ablation of the value of $k$ is in the 
\supp
Section~\ref{sec:clusters_number_ablation}). To extract segment-wise features used for k-means clustering, we use \texttt{CLS} token features of the DINO-trained~\cite{caron2021emerging} ViT-S~\cite{dosovitskiy2021image} model. The ablation of the feature extractor is in Section~\ref{sec:ablations}. Obtained segment-wise labels serve as pseudo-annotations for training the ResNet18-FPN and Segmenter models, as discussed in Section~\ref{sec:clustering}.

\emph{Pre-training data and networks.}
To be directly comparable to~\cite{cho2021picie}, in our experiments with the ResNet18+FPN model, we initialize its backbone with a ResNet18 trained with supervision on the ImageNet-1k~\cite{deng2009imagenet} classification task, exactly as all the compared prior methods (PiCIE, DC, and IIC). However, we aim at having a completely unsupervised setup. Therefore, in our experiments with Segmenter, we initialize its backbone with a ViT-S also trained on ImageNet-1k~\cite{deng2009imagenet} but with the self-supervised DINO approach~\cite{caron2021emerging}. %
Please note that for both backbones we use the same pre-training data (ImageNet-1k) exactly as the prior methods PiCIE, DC, and IIC with which we compare.
These pre-trained backbones are then used as an initialization for training the actual segmentation models on specific autonomous driving datasets as described next.

\emph{Training datasets.}
We train our models on about 7k images from the Waymo Open~\cite{sun2020scalability} dataset, which has both image and LiDAR data available.
For the baseline methods (IIC~\cite{ji2019invariant}, modified DC~\cite{caron2018deep} and PiCIE~\cite{cho2021picie}) we follow the setup from~\cite{cho2021picie}, i.e., we train the models on all available images of Cityscapes~\cite{Cordts2016Cityscapes}, meaning the $24.5$k images from the \textit{train}, \textit{test}, and \textit{train\_extra} splits. Note that those models then do not face the problem of domain gap when evaluated on the Cityscapes~\cite{Cordts2016Cityscapes} dataset.
To be directly comparable with our approach, we also train a variant of modified DC~\cite{caron2018deep} and PiCIE~\cite{cho2021picie} on the same subset of the Waymo Open~\cite{sun2020scalability} dataset as used in our approach. 
Furthermore, to test the generalizability of our method to other training datasets, we provide results of modified DC, PiCIE, and our \ours approach when trained on the nuScenes~\cite{nuscenes} dataset in Sec.~\ref{sec:nuscenes_results} in the 
\supp.

\emph{Optimization.} To train IIC~\cite{ji2019invariant}, modified DC~\cite{caron2018deep}, and PiCIE~\cite{cho2021picie}, we use the setup provided in~\cite{cho2021picie}.
For our \ours, we train the teacher and student models 
with batches of size $32$ and with a learning rate of $2e{-}4$ with a polynomial schedule on a single 
V100 GPU. 
During training, we perform data augmentation consisting of random image resizing in the $\left(0.5, 2.0\right)$ range, random cropping with the crop size of $512\times 512$ pixels, random horizontal flipping, and photometric distortions.%

\paragraphcustom{Evaluation protocol.}
\label{sec:evaluation_protocol}
\emph{Mapping.} To evaluate our models in the unsupervised setup, we run trained models on every image, thus getting segmentation predictions with values from 1 to $k$.  Then, we compute the confusion matrix between the $C$ ground-truth classes of the target dataset and the $k\ge C$ pseudo-classes. We map the $C$ ground-truth classes to $C$ out of the $k$ pseudo-classes using Hungarian matching~\cite{Kuhn1955thehungarian} that minimizes the overall confusion. Finally, we compute the 
mean IoU and pixel accuracy based on this mapping. The pixel predictions for the $k-C$ unmapped pseudo-classes are considered as false negatives.

\emph{Test datasets.}
We evaluate the trained models on Cityscapes~\cite{Cordts2016Cityscapes},  Dark Zurich~\cite{SDV20}, Nighttime driving~\cite{daytime:2:nighttime} and ACDC~\cite{SDV21} datasets in this fully unsupervised setup\footnote{ These adverse weather datasets~\cite{daytime:2:nighttime,SDV20,SDV21} are commonly used by domain adaptation approaches that leverage unlabeled images of this type for adaptation. Here, we consider them only for evaluation to assess the generalization of our strategy; we do not have access to any of those images during training.} \emph{without any finetuning}. Cityscapes~\cite{Cordts2016Cityscapes} is a well-established dataset 
with 500 validation images that we use for evaluation. %
Dark Zurich~\cite{SDV20} and Nighttime driving~\cite{daytime:2:nighttime} are two nighttime datasets, each with $50$ validation images annotated for semantic segmentation that we use for evaluation. ACDC~\cite{SDV21} is a recent dataset providing four different adverse weather conditions with $400$ training and $100$ validation samples per weather condition. We test our approach on the validation images annotated for semantic segmentation. The Cityscapes dataset defines $30$ different semantic classes for the pixel-wise semantic segmentation task. We follow prior work, and, unless stated otherwise, we evaluate our approach on the pre-defined subset of $19$ classes \cite{Cordts2016Cityscapes}. The same set of $19$ classes is used for all other datasets.

\emph{Metrics.} We evaluate results with two standard metrics for the semantic segmentation task, the mean Intersection over Union, \emph{mIoU}, and the pixel accuracy, \emph{PA}, as done in prior work~\cite{cho2021picie}. The mIoU is the mean intersection over union averaged over all classes, while PA defines the percentage of pixels in the image that are segmented correctly, averaged over all images.

\vspace{-10pt}
\subsection{Comparison to state of the art}

Here we evaluate our trained models in the unsupervised setup using the evaluation protocol described in Section~\ref{sec:evaluation_protocol}. 
We compare our method using both the ResNet18+FPN and Segmenter~\cite{Strudel_2021_ICCV} models to three recent unsupervised segmentation models: IIC~\cite{ji2019invariant}, modified version of DeepCluster~\cite{caron2018deep} (DC), and PiCIE~\cite{cho2021picie}. In the 
\supp,
we assess the utility of the features learned by our model in other settings, such as linear probing for semantic segmentation (Sec.~\ref{sec:fine}), and k-NN evaluation (Sec.~\ref{sec:knn_eval}).

\begingroup
\begin{table*}[t!]
\scriptsize
\centering
\caption{
    \textbf{Comparison to the state of the art} for unsupervised semantic segmentation on Cityscapes~\cite{Cordts2016Cityscapes} (CS), DarkZurich~\cite{SDV20} (DZ) and Nighttime driving~\cite{daytime:2:nighttime} (ND) datasets measured by the mean IoU (mIoU). 
    The colored differences are reported with respect to the state-of-the-art approach of~\cite{cho2021picie} denoted by 
    {\tiny \faAnchor}. 
    The \emph{sup. init.} abbreviation stands for supervised initialization of the \emph{encoder}, and the column
    \emph{train. data} indicates whether Cityscapes (CS) or Waymo Open (WO) dataset was used for training.
    Please see the appendix for pixel accuracy and for results using the nuScenes dataset for training.
}
\label{table:zeroshot_eval}
\vspace{-5pt}
\begin{tabular}{lcc|S[table-format=2.1]r|
S[table-format=2.1]r|
S[table-format=2.1]r|
S[table-format=2.1]r}
\bottomrule
 & sup. & train. &  \multicolumn{2}{c|}{CS19~\cite{Cordts2016Cityscapes}} & \multicolumn{2}{c|}{CS27~\cite{Cordts2016Cityscapes}} & \multicolumn{2}{c|}{DZ~\cite{SDV20}} & \multicolumn{2}{c}{ND~\cite{daytime:2:nighttime}} \\
\textcolor{gray}{architecture,} method & init. & data &
\multicolumn{2}{c|}{mIoU} & 
\multicolumn{2}{c|}{mIoU} & 
\multicolumn{2}{c|}{mIoU} &
\multicolumn{2}{c}{mIoU} \\
\hline\hline
\textcolor{gray}{RN18+FPN}  & & & & & & & & & & \\
\rowcolor{Gray!10!white}\hspace{3mm}IIC$^\dagger$~\cite{ji2019invariant} & yes & CS & \text{-} & & 6.4 &  {(\Minus4.8)} & \text{-} & & \text{-} & \\
\rowcolor{Gray!10!white}\hspace{3mm}Modified DC$^\ddagger$~\cite{caron2018deep} & yes &  CS & 11.3 &  {(\Minus4.5)} & 
7.9 &  {(\Minus3.3)} & 
7.5 &  {\color{better}{(\Plus2.9)}} &  
8.2 &  {(\Minus1.3)}
\\
\rowcolor{Gray!10!white}{\tiny \faAnchor}\hspace{1mm}PiCIE$^\ddagger$~\cite{cho2021picie} & yes &  CS &
15.8 & & 11.2 & & 4.6 & & 9.5 &  \\

\hspace{3mm}Modified DC$^\star$ & yes & WO & 11.4 &  {(\Minus4.4)} & 7.0 &  {(\Minus4.1)} & 5.9 & \color{better}{ {(\Plus1.3)}} & 8.2 &  {(\Minus1.3)} \\
\hspace{3mm}PiCIE$^\star$ & yes & WO & 
13.7 &  {(\Minus2.1)} & 
9.7 &  {(\Minus1.5)} & 
4.9 &  {\color{better}{(\Plus0.3)}} & 
9.3 &  {(\Minus0.2)} \\
\hspace{3mm}\ours (Ours, $\student$) 
& yes &  WO &
19.5 &  {\color{better}{(\Plus3.7)}} &
\textbf{16.2} &  {\color{better}{(\textbf{\Plus5.1})}} & 
10.9 &  {\color{better}{(\Plus6.3)}} & 
14.4 &  {\color{better}{(\Plus4.9)}} \\

\hline
\multicolumn{3}{l|}{\textcolor{gray}{Segmenter, ViT-S/16}} & & & & & & & & \\

\hspace{3mm}\ours (Ours, $\student$) 
& no & WO &
\textbf{21.8} &  {\color{better}{(\textbf{\Plus6.0})}} &
15.3 &  {\color{better}{(\Plus4.1)}} & 
\textbf{14.2} &  {\color{better}{(\textbf{\Plus9.6})}} &
\textbf{18.9} &  {\color{better}{(\textbf{\Plus9.3})}} \\
\toprule
\multicolumn{11}{l}{\scriptsize \textsuperscript{$\dagger$} Results reported in~\cite{cho2021picie}. \textsuperscript{$\ddagger$} Models provided by the PiCIE~\cite{cho2021picie} authors.} \\
\multicolumn{11}{l}{\textsuperscript{$*$} Trained by PiCIE code base.}
\end{tabular}
\end{table*}
\endgroup

\begingroup
\setlength{\tabcolsep}{0.6pt} 
\begin{table}[t!]
\centering
\caption{\textbf{Comparison to the state-of-the-art} %
for unsupervised semantic segmentation on the ACDC~\cite{SDV21} dataset. 
Please refer to Table~\ref{table:zeroshot_eval} for the used symbols.
Please see the appendix for pixel accuracy and for results using the nuScenes dataset for training.
}
\vspace{-5pt}
\label{table:zeroshot_acdc}
\scriptsize
\begin{tabular}{lcc|
S[table-format=2.1]r|
S[table-format=2.1]r|
S[table-format=2.1]r|
S[table-format=2.1]r||
S[table-format=2.1]r}
\bottomrule
& sup. & train. & \multicolumn{2}{c|}{night} & \multicolumn{2}{c|}{fog} & \multicolumn{2}{c|}{rain} & \multicolumn{2}{c||}{snow}& \multicolumn{2}{c}{average} \\
\textcolor{gray}{architecture,} method & init. & data &
\multicolumn{2}{c|}{mIoU} &
\multicolumn{2}{c|}{mIoU} &
\multicolumn{2}{c|}{mIoU} &
\multicolumn{2}{c||}{mIoU} &
\multicolumn{2}{c}{mIoU} \\
\hline\hline
\textcolor{gray}{RN18+FPN} & & & & & & & & & & & & \\
\rowcolor{Gray!10!white}\hspace{2.5mm}modified DC$^\ddagger$~\cite{caron2018deep}& yes & CS &
8.1 & {\color{better}{(\Plus3.7})} & 
8.3 & {(\Minus4.0)} &
6.9 & {(\Minus5.6)} &
7.4 & {(\Minus4.7)} &
7.7 & {(\Minus2.6)} \\
\rowcolor{Gray!10!white}{\tiny \faAnchor}\hspace{0.5mm}PiCIE$^\ddagger$~\cite{cho2021picie}& yes & CS & 4.4 & & 12.2 & & 12.5 & & 12.1 & & 10.3 & \\

\hspace{2.5mm}modified DC$^\star$ & yes & WO &
5.9 & \color{better}{{(\Plus1.5)}} & 
11.7 & {(\Minus0.5)} & 9.6 & {(\Minus2.9)} & 9.8 & {(\Minus2.3)} & 9.2 & {(\Minus1.0)} \\
\hspace{2.5mm}PiCIE$^\star$ & yes & WO &
4.7 & \color{better}{{(\Plus0.3)}} & 14.4 & \color{better}{{(\Plus2.1)}} & 13.7 & \color{better}{{(\Plus1.2)}} &  14.3 & \color{better}{{(\Plus2.2)}} & 11.7 & \color{better}{{(\Plus1.5)}} \\
\hspace{2.5mm}\ours (Ours, $\student$) 
& yes & WO &
11.2 & {\color{better}{(\Plus6.8)}} & 
14.5 & {\color{better}{(\Plus2.3)}} & 
14.9 & {\color{better}{(\Plus2.5)}} & 
14.6 & {\color{better}{(\Plus2.6)}} & 
13.8 & {\color{better}{(\Plus3.5)}} \\
\hline
\multicolumn{3}{l|}{\textcolor{gray}{Segmenter, ViT-S/16}} & & & & &  & & & & & \\

\ifnuscenes
\hspace{2.5mm}\AV{\ours (Ours, $\teacher$)} & no & \AV{nuSC} &
11.3 & \scriptsize{\color{better}{\textbf{}}} &
12.2 & \scriptsize{\color{better}{\textbf{}}} & 
13.5 & \scriptsize{\color{better}{\textbf{}}} & 
13.0 & \scriptsize{\color{better}{\textbf{}}} & 
12.5 & \scriptsize{\color{better}{\textbf{}}}
\\
\hspace{2.5mm}\AV{\ours (Ours, $\student$)} & no & \AV{nuSC} &
10.6 & \scriptsize{\color{better}{\textbf{}}} & 
13.3 & \scriptsize{\color{better}{\textbf{}}} & 
16.0 & \scriptsize{\color{better}{\textbf{}}} & 
14.8 & \scriptsize{\color{better}{\textbf{}}} & 
13.9 & \scriptsize{\color{better}{\textbf{}}}
\\
\fi

\hspace{2.5mm}\ours (Ours, $\student$) 
& no & WO &
\bf{13.8} & {\color{better}{(\textbf{\Plus9.4})}} & 
\bf{18.1} & {\color{better}{(\textbf{\Plus5.9})}} & 
\bf{16.4} & {\color{better}{(\textbf{\Plus3.9})}} & 
\bf{18.7} & {\color{better}{(\textbf{\Plus6.6})}} & 
\bf{16.7} & {\color{better}{(\textbf{\Plus6.5})}} \\
\hline
\end{tabular}
\end{table}
\endgroup
We provide results on the Cityscapes, Dark Zurich, and Nighttime Driving datasets in Table~\ref{table:zeroshot_eval}, and show qualitative results in Figure~\ref{fig:qualitative}.
As shown in the first two columns of Table~\ref{table:zeroshot_eval},
our approach outperforms~\cite{cho2021picie} on the Cityscapes dataset by a large margin
 in both the $19$-class and $27$-class set-ups.
Improvements are visible for both architectures, but in most cases, the best results are obtained with the distilled Segmenter architecture using the ViT-S/16 backbone.
The last two columns in Table~\ref{table:zeroshot_eval} show results for the two nighttime segmentation datasets: Dark Zurich~\cite{SDV20} and Nighttime Driving~\cite{daytime:2:nighttime}. Our models again outperform~\cite{cho2021picie} in all setups. In addition, we observe a 
better performance of our models compared to~\cite{cho2021picie} when evaluating on the nighttime scenes.
For example, on the Dark Zurich~\cite{SDV20} dataset, the mIoU of PiCIE~\cite{cho2021picie} decreases by $71\%$ compared to the results on Cityscapes ($15.8\rightarrow 4.6$), while the mIoU of our 
Segmenter-based model decreases only by $35\%$ ($21.8\rightarrow 14.2$). This suggests that our models generalize significantly better to out-of-distribution scenes.
These findings hold for PiCIE models trained on both Cityscapes and on Waymo Open Dataset.

Finally, Table~\ref{table:zeroshot_acdc} shows results on the ACDC dataset in four different adverse weather conditions. Results follow a similar trend as in Table~\ref{table:zeroshot_eval} and show the superiority of our approach measured by mIoU compared
to the current state-of-the-art unsupervised semantic segmentation method of~\cite{cho2021picie} on images out of the training distribution, such as images at night or in snow. Please see the 
\supp
for the complete set of results, including results using nuScenes (Sec.~\ref{sec:nuscenes_results}), pixel accuracy (Sec.~\ref{sec:pixel_accuracy}), per-class results (Sec.~\ref{sec:per_class_results}) and analysis of the confusion matrices (Sec.~\ref{sec:confusion}).

\vspace{-5pt}
\subsection{Ablations}
\label{sec:ablations}

\begin{table}[t!]
    \caption{
      {\bf Ablations on the Cityscapes dataset.}
      \textbf{(a)} Benefits of our segment extraction method over segment proposals from~\cite{felzenszwalb2004efficient}.
      \textbf{(b)} Benefits of our distillation approach showing an improvement of the student~($\student$) over the the teacher~($\teacher$) and benefits of our LiDAR cross-modal spatial constraints (LiD).
      \textbf{(c)} Ablation of different feature extractors for the k-means clustering.
    }
    \addtocounter{table}{-1} %
    \vspace{-15pt}
    \centering
    \subfloat[][Segment extraction \label{table:segmentation_comparison}]{%
        \resizebox{0.3\columnwidth}{!}{%
        \scriptsize
\begin{tabular}[t]{l|rrrr}
\bottomrule
\textcolor{gray}{arch.} & & & & \\
\hspace{1mm}seg. prop. & \multicolumn{2}{c}{mIoU} & \multicolumn{2}{c}{PA} \\
\hline
\textcolor{gray}{RN18\Plus FPN} & & & & \\
\hspace{1mm}FH~\cite{felzenszwalb2004efficient} & 15.5 &  & 52.8 & \\
\hspace{1mm}Ours & \textbf{17.4} & \tiny{\color{better}{\textbf{(\Plus1.9)}}} &
\textbf{55.9} & \tiny{\color{better}{\textbf{(\Plus3.1)}}} \\
\hline
\textcolor{gray}{Segmenter} & & & & \\
\hspace{1mm}FH~\cite{felzenszwalb2004efficient} &  15.8 & & 51.8 & \\
\hspace{1mm}Ours & 
\textbf{20.4} & \tiny{\color{better}{\textbf{(\Plus4.6)}}}  &
\textbf{65.4} & \tiny{\color{better}{\textbf{(\Plus13.6)}}}  \\
\toprule
\end{tabular}

        }
    }%
    \subfloat[][Distillation\label{table:distillation_comparison}]{%
        \resizebox{0.355\columnwidth}{!}{%
        \scriptsize
\begin{tabular}[t]{lc|S[table-format=2.1]rS[table-format=2.1]r}
    \bottomrule
    model & LiD. & \multicolumn{2}{c}{mIoU} & \multicolumn{2}{c}{PA} \\
    \hline
    \multicolumn{2}{l|}{\textcolor{gray}{RN18\Plus FPN}} & & & \\
    \rowcolor{Gray!10!white}PiCIE~($\teacher$) & & 13.7 & & 48.6 &  \\
    \rowcolor{Gray!10!white}PiCIE~($\student$) & & 14.8 & \tiny{\color{better}{(\Plus1.1)}} & 64.1 &  \tiny{\color{better}{(\Plus15.5)}} \\ %
    \rowcolor{Gray!10!white}PiCIE~($\student$) & \checkmark & 15.1 & \tiny{\color{better}{(\Plus1.4)}} & \textbf{68.4}  & \tiny{\color{better}{\textbf{(\Plus19.8)}}} \\
    Ours~($\teacher$) & & 17.4 &  & 55.9 & \\
    Ours~($\student$) & & 18.8 & \tiny{\color{better}{(\Plus1.4)}} &
    63.4 & \tiny{\color{better}{(\Plus7.5)}} \\ %
    Ours~($\student$) & \checkmark & \textbf{19.5} & \tiny{\color{better}{(\textbf{\Plus2.1})}} &
    \textbf{66.4} & \tiny{\color{better}{(\textbf{\Plus10.5})}} \\
    \hline
    \multicolumn{2}{l|}{\textcolor{gray}{Segmenter}} & & & \\
    Ours~($\teacher$) & & 20.4 & & 65.4 & \\
    Ours~($\student$) & &
    20.8 & \tiny{\color{better}{(\Plus0.4)}}  &
    68.5 & \tiny{\color{better}{(\Plus3.1)}}  \\ %
    Ours~($\student$) & \checkmark & %
    \textbf{21.8} & \tiny{\color{better}{(\textbf{\Plus1.4})}}  &
    \textbf{69.5} & \tiny{\color{better}{(\textbf{\Plus4.1})}}  \\
    \toprule
\end{tabular}

        }
    }%
    \subfloat[][Feature extractors\label{table:feature_extractors}]{%
        \resizebox{0.261\columnwidth}{!}{%
        \scriptsize
\begin{tabular}[t]{l|rr}
\bottomrule
\textcolor{gray}{arch.} & & \\
\hspace{1mm}method & mIoU & PA \\
\hline
\textcolor{gray}{ViT-S/16} & & \\
\hspace{1mm}DeiT~\cite{touvron2021training} & 21.7 & 73.0 \\
\hspace{1mm}DINO~\cite{caron2021emerging} & 20.2 & 64.4 \\
\hline
\textcolor{gray}{ResNet18} & & \\
\hspace{1mm}supervised~\cite{he2016deep} & 19.6 & 70.0 \\
\hline
\textcolor{gray}{ResNet50} & &  \\
\hspace{1mm}supervised~\cite{he2016deep} & 21.3 & 67.6 \\
\hspace{1mm}OBOW~\cite{gidaris2021obow} & 20.7 & 65.9 \\
\hspace{1mm}PixPro~\cite{xie2020propagate} & 20.7 & 65.9 \\
\hspace{1mm}MaskCon.~\cite{van2021unsupervised} & 19.1 & 68.0 \\
\toprule
\end{tabular}

        }
    }
    \vspace{-15pt}
    \addtocounter{table}{+1}
\end{table}

In this section, we ablate the main components of our 
approach. In particular, we study %
the benefits of our cross-modal segment extraction (Table~\ref{table:segmentation_comparison}),
of our distillation with cross-modal spatial constraints (Table~\ref{table:distillation_comparison}), 
effect of varying feature extractors for the k-means clustering (Table~\ref{table:feature_extractors}),
variance of the results over multiple runs, the influence of LiDAR resolution and the number of clusters used in k-means.

\paragraphcustom{Segment extraction approach.}
To evaluate the benefits of our cross-modal segment extraction module, we investigate using segment proposals generated with a purely image-based segmentation approach by Felzenszwalb and Huttenlocher (FH)~\cite{felzenszwalb2004efficient}. It groups pixels into segments based on similar color and texture properties.
We use the same set of hyperparameters as~\cite{henaff2021efficient}.  
The results are shown in Table~\ref{table:segmentation_comparison} 
and demonstrate clear benefits of our LiDAR-based cross-modal segment extraction method despite the difficulties of using LiDAR data discussed in Section~\ref{sec:segment_extraction}. 
We attribute the better results of our approach
to the fact that LiDAR 
data
segmentation operates with range information, which is much stronger at separating objects from the background and from each other compared to the purely image-based approach of FH~\cite{felzenszwalb2004efficient}. Indeed, FH relies only on color/texture and is therefore much more likely to join multiple objects into one segment or separate a single object into multiple segments. 
The benefits of our 
cross-modal segment extraction 
are observed for 
both studied
architectures.

\paragraphcustom{Distillation with cross-modal spatial constraints.} 
To evaluate the benefits of our teacher-student distillation method with cross-modal spatial constraints (Section~\ref{sec:distillation}), we compare the predictions of the teacher~$\teacher $ (before distillation) and the student~$\student $ (after distillation).
Table~\ref{table:distillation_comparison} presents results on the Cityscapes dataset using both convolutional- and transformer-based architectures.
The results show consistent improvements using our distillation technique, particularly regarding the pixel accuracy metric. 
We believe that this could be attributed to improvements in predictions for classes such as vegetation and buildings. They often occupy large areas of the image and benefit most from the distillation as they are usually not well covered by the LiDAR scans. 
Furthermore, the results show clear benefits of using this distillation step both with and without cross-modal spatial constraints (LiD) by Student~$\student$ outperforming Teacher~$\teacher$ in both scenarios. 
Please also note that our distillation technique works well even in combination with another training approach (PiCIE~\cite{cho2021picie}).

\paragraphcustom{Sensitivity to the initialization.}
To study the influence of initialization, we take the features extracted by DINO~\cite{caron2021emerging} and run the k-means clustering (Section~\ref{sec:clustering}) four times. 
For each of the k-means clustering outcomes, we run the segmentation model training four times with different initializations.
The variance over all k-means and training runs is only $0.5$ for mIoU and $1.5$ for pixel accuracy (i.e., $20.4\pm0.5 / 65.4\pm1.5$).
These results clearly show that our method is not very sensitive to k-means initialization or to the network initialization.

\paragraphcustom{Feature extractors.} 
An ablation of seven different feature extractors (convolutional- and Transformer-based) for the task of segment-wise unsupervised labelling is shown in Table~\ref{table:feature_extractors}. 
The results on the Cityscapes~\cite{Cordts2016Cityscapes} dataset using our Segmenter model demonstrate that our approach works well with several different feature extractors.

\paragraphcustom{LiDAR resolution and number of clusters.}
An ablation of the influence of LiDAR resolution is shown in Sec.~\ref{sec:lidar_density_ablation} in the 
\supp
and demonstrates that our method is robust to LiDAR’s sparsity.
Furthermore, we study the choice of the number of clusters for the k-means clustering in Sec.~\ref{sec:clusters_number_ablation} in the 
\supp.

\vspace{-5pt}
\subsection{Limitations and failure modes}
Our approach has the following three main limitations.
First, LiDAR point clouds do not provide information about very distant or even infinitely distant objects, e.g., the sky, which our approach cannot learn to segment.
Second, LiDAR point clouds paired with geometric segmentation %
can not correctly distinguish road from sidewalk or grass, when all surfaces are similarly flat. Both the above limitations might be possibly tackled by pairing our LiDAR-based segment proposals with an unsupervised image-based method such as~\cite{felzenszwalb2004efficient}. 
Also, the LiDAR points must not be too sparse (e.g., 
only 4 beams), since otherwise the LiDAR-based segments would be of poor quality. However, this is not an overly restricting requirement as it is common 
to use LiDAR sensors with sufficient beam resolution, e.g., as in the recent Waymo Open~\cite{sun2020scalability} or ONCE~\cite{mao2021one} datasets.
Finally, we encounter semantically similar objects appearing in multiple pseudo-classes, a natural side effect of clustering. This issue may be mitigated by using different feature clustering methods, e.g., from the family of graph clustering methods, that allow the measurement of similarities on manifolds in the feature space instead of the currently used Euclidean metric in the k-means clustering.

\vspace{-8pt}
\section{Conclusion}
We have developed \ours - a fully unsupervised approach for semantic image segmentation in urban scenes. The approach relies on novel modules for (i) cross-modal segment extraction and (ii) distillation with cross-modal constraints that leverage LiDAR point clouds aligned with images. We evaluate our approach on four different autonomous driving datasets in challenging weather and illumination conditions and demonstrate major gains over prior work. This work opens up the possibility of large-scale autonomous learning of embodied perception models without explicit human supervision.

\subsubsection{Acknowledgments.}
This work was partly supported by the European Regional Development Fund under the project IMPACT (reg. no. CZ.02.1.01\/0.0\/0.0\/15\_003\/0000468), and the Ministry of Education, Youth and Sports of the Czech Republic through the e-INFRA CZ (ID:90140). Antonin Vobecky was partly supported by the CTU Student Grant Agency (reg. no. SGS21\/184\/OHK3\/3T\/37).

\bibliographystyle{splncs04}
\bibliography{main}

\begin{thebibliography}{10}
\providecommand{\url}[1]{\texttt{#1}}
\providecommand{\urlprefix}{URL }
\providecommand{\doi}[1]{https://doi.org/#1}

\bibitem{afouras2021self}
Afouras, T., Asano, Y.M., Fagan, F., Vedaldi, A., Metze, F.: Self-supervised
  object detection from audio-visual correspondence. In: arXiv (2021)

\bibitem{alayrac2020self}
Alayrac, J.B., Recasens, A., Schneider, R., Arandjelovic, R., Ramapuram, J.,
  De~Fauw, J., Smaira, L., Dieleman, S., Zisserman, A.: Self-supervised
  multimodal versatile networks. In: NeurIPS (2020)

\bibitem{alwassel_2020_xdc}
Alwassel, H., Mahajan, D., Korbar, B., Torresani, L., Ghanem, B., Tran, D.:
  Self-supervised learning by cross-modal audio-video clustering. In: NeurIPS
  (2020)

\bibitem{arandjelovic2017look}
Arandjelovic, R., Zisserman, A.: Look, listen and learn. In: ICCV (2017)

\bibitem{bartoccioni2021lidartouch}
Bartoccioni, F., Zablocki, {\'E}., P{\'e}rez, P., Cord, M., Alahari, K.:
  Lidartouch: Monocular metric depth estimation with a few-beam lidar. In:
  arXiv (2021)

\bibitem{bielski2019emergence}
Bielski, A., Favaro, P.: Emergence of object segmentation in perturbed
  generative models. In: NeurIPS (2019)

\bibitem{bogoslavskyi17pfg}
Bogoslavskyi, I., Stachniss, C.: Efficient online segmentation for sparse 3d
  laser scans. PFG  (2017)

\bibitem{nuscenes}
Caesar, H., Bankiti, V., Lang, A.H., Vora, S., Liong, V.E., Xu, Q., Krishnan,
  A., Pan, Y., Baldan, G., Beijbom, O.: nuscenes: A multimodal dataset for
  autonomous driving. In: CVPR (2020)

\bibitem{caron2018deep}
Caron, M., Bojanowski, P., Joulin, A., Douze, M.: Deep clustering for
  unsupervised learning of visual features. In: {ECCV} (2018)

\bibitem{caron2021emerging}
Caron, M., Touvron, H., Misra, I., Jegou, H., Mairal, J., Bojanowski, P.,
  Joulin, A.: {Emerging Properties in Self-Supervised Vision Transformers}. In:
  ICCV (2021)

\bibitem{chen2021localizing}
Chen, H., Xie, W., Afouras, T., Nagrani, A., Vedaldi, A., Zisserman, A.:
  Localizing visual sounds the hard way. In: CVPR (2021)

\bibitem{chen2018encoder}
Chen, L.C., Zhu, Y., Papandreou, G., Schroff, F., Adam, H.: Encoder-decoder
  with atrous separable convolution for semantic image segmentation. In: ECCV
  (2018)

\bibitem{chen2019unsupervised}
Chen, M., Arti{\`e}res, T., Denoyer, L.: Unsupervised object segmentation by
  redrawing. In: NeurIPS (2019)

\bibitem{Cheng_2020_CVPR}
Cheng, B., Collins, M.D., Zhu, Y., Liu, T., Huang, T.S., Adam, H., Chen, L.C.:
  Panoptic-deeplab: A simple, strong, and fast baseline for bottom-up panoptic
  segmentation. In: CVPR (2020)

\bibitem{cho2021picie}
Cho, J.H., Mall, U., Bala, K., Hariharan, B.: {PiCIE}: Unsupervised semantic
  segmentation using invariance and equivariance in clustering. In: CVPR (2021)

\bibitem{Cordts2016Cityscapes}
Cordts, M., Omran, M., Ramos, S., Rehfeld, T., Enzweiler, M., Benenson, R.,
  Franke, U., Roth, S., Schiele, B.: The cityscapes dataset for semantic urban
  scene understanding. In: CVPR (2016)

\bibitem{daytime:2:nighttime}
Dai, D., {Van Gool}, L.: Dark model adaptation: Semantic image segmentation
  from daytime to nighttime. In: IEEE ITSC (2018)

\bibitem{deng2009imagenet}
Deng, J., Dong, W., Socher, R., Li, L.J., Li, K., Fei-Fei, L.: {ImageNet: A
  Large-Scale Hierarchical Image Database}. In: CVPR (2009)

\bibitem{dosovitskiy2021image}
Dosovitskiy, A., Beyer, L., Kolesnikov, A., Weissenborn, D., Zhai, X.,
  Unterthiner, T., Dehghani, M., Minderer, M., Heigold, G., Gelly, S.,
  Uszkoreit, J., Houlsby, N.: An image is worth 16x16 words: Transformers for
  image recognition at scale. In: ICLR (2021)

\bibitem{felzenszwalb2004efficient}
Felzenszwalb, P.F., Huttenlocher, D.P.: Efficient graph-based image
  segmentation. IJCV  (2004)

\bibitem{french2020semi}
French, G., Laine, S., Aila, T., Mackiewicz, M., Finlayson, G.: Semi-supervised
  semantic segmentation needs strong, varied perturbations. BMVC  (2020)

\bibitem{gidaris2021obow}
Gidaris, S., Bursuc, A., Puy, G., Komodakis, N., Cord, M., P{\'{e}}rez, P.:
  Obow: Online bag-of-visual-words generation for self-supervised learning. In:
  CVPR (2021)

\bibitem{Gidaris2018Unsupervised}
Gidaris, S., Singh, P., Komodakis, N.: Unsupervised representation learning by
  predicting image rotations. In: ICLR (2018)

\bibitem{grill2020bootstrap}
Grill, J., Strub, F., Altch{\'{e}}, F., Tallec, C., Richemond, P.H.,
  Buchatskaya, E., Doersch, C., Pires, B.{\'{A}}., Guo, Z., Azar, M.G., Piot,
  B., Kavukcuoglu, K., Munos, R., Valko, M.: Bootstrap your own latent - {A}
  new approach to self-supervised learning. In: NeurIPS (2020)

\bibitem{he2020momentum}
He, K., Fan, H., Wu, Y., Xie, S., Girshick, R.B.: Momentum contrast for
  unsupervised visual representation learning. In: CVPR (2020)

\bibitem{he2016deep}
He, K., Zhang, X., Ren, S., Sun, J.: Deep residual learning for image
  recognition. In: CVPR (2016)

\bibitem{henaff2021efficient}
H{\'e}naff, O.J., Koppula, S., Alayrac, J.B., Oord, A.v.d., Vinyals, O.,
  Carreira, J.: Efficient visual pretraining with contrastive detection. In:
  ICCV (2021)

\bibitem{hwang2019segsort}
Hwang, J.J., Yu, S.X., Shi, J., Collins, M.D., Yang, T.J., Zhang, X., Chen,
  L.C.: Segsort: Segmentation by discriminative sorting of segments. In: ICCV.
  pp. 7334--7344 (2019)

\bibitem{jaritz2020xmuda}
Jaritz, M., Vu, T.H., Charette, R.d., Wirbel, E., P{\'e}rez, P.: xmuda:
  Cross-modal unsupervised domain adaptation for 3d semantic segmentation. In:
  CVPR (2020)

\bibitem{ji2019invariant}
Ji, X., Henriques, J.F., Vedaldi, A.: Invariant information clustering for
  unsupervised image classification and segmentation. In: ICCV (2019)

\bibitem{kanezaki2018unsupervised}
Kanezaki, A.: Unsupervised image segmentation by backpropagation. In: ICASSP
  (2018)

\bibitem{Kuhn1955thehungarian}
Kuhn, H.W., Yaw, B.: The hungarian method for the assignment problem. NRLQ
  (1955)

\bibitem{lin2017feature}
Lin, T.Y., Doll{\'a}r, P., Girshick, R., He, K., Hariharan, B., Belongie, S.:
  Feature pyramid networks for object detection. In: CVPR (2017)

\bibitem{long2015fully}
Long, J., Shelhamer, E., Darrell, T.: Fully convolutional networks for semantic
  segmentation. In: CVPR (2015)

\bibitem{mao2021one}
Mao, J., Niu, M., Jiang, C., Liang, H., Liang, X., Li, Y., Ye, C., Zhang, W.,
  Li, Z., Yu, J., et~al.: One million scenes for autonomous driving: Once
  dataset. NeurIPS  (2021)

\bibitem{miech2020end}
Miech, A., Alayrac, J.B., Smaira, L., Laptev, I., Sivic, J., Zisserman, A.:
  End-to-end learning of visual representations from uncurated instructional
  videos. In: CVPR (2020)

\bibitem{neuhold2017mapillary}
Neuhold, G., Ollmann, T., Rota~Bulo, S., Kontschieder, P.: The mapillary vistas
  dataset for semantic understanding of street scenes. In: ICCV (2017)

\bibitem{ouali2020autoregressive}
Ouali, Y., Hudelot, C., Tami, M.: Autoregressive unsupervised image
  segmentation. In: ECCV (2020)

\bibitem{owens2018audio}
Owens, A., Efros, A.A.: Audio-visual scene analysis with self-supervised
  multisensory features. In: ECCV (2018)

\bibitem{Recasens_2021_ICCV}
Recasens, A., Luc, P., Alayrac, J.B., Wang, L., Strub, F., Tallec, C.,
  Malinowski, M., P\u{a}tr\u{a}ucean, V., Altch\'e, F., Valko, M., Grill, J.B.,
  van~den Oord, A., Zisserman, A.: Broaden your views for self-supervised video
  learning. In: ICCV (2021)

\bibitem{Ronneberger2015UNet}
Ronneberger, O., Fischer, P., Brox, T.: U-net: Convolutional networks for
  biomedical image segmentation. In: MICCAI (2015)

\bibitem{SDV20}
Sakaridis, C., Dai, D., Van~Gool, L.: Map-guided curriculum domain adaptation
  and uncertainty-aware evaluation for semantic nighttime image segmentation.
  IEEE TPAMI  (2020)

\bibitem{SDV21}
Sakaridis, C., Dai, D., Van~Gool, L.: {ACDC}: The adverse conditions dataset
  with correspondences for semantic driving scene understanding. In: ICCV
  (2021)

\bibitem{Strudel_2021_ICCV}
Strudel, R., Garcia, R., Laptev, I., Schmid, C.: Segmenter: Transformer for
  semantic segmentation. In: ICCV (2021)

\bibitem{sun2020scalability}
Sun, P., Kretzschmar, H., Dotiwalla, X., Chouard, A., Patnaik, V., Tsui, P.,
  Guo, J., Zhou, Y., Chai, Y., Caine, B., et~al.: Scalability in perception for
  autonomous driving: Waymo open dataset. In: CVPR (2020)

\bibitem{tian2021unsupervised}
Tian, H., Chen, Y., Dai, J., Zhang, Z., Zhu, X.: Unsupervised object detection
  with lidar clues. In: CVPR (2021)

\bibitem{touvron2021training}
Touvron, H., Cord, M., Douze, M., Massa, F., Sablayrolles, A., J{\'e}gou, H.:
  Training data-efficient image transformers \& distillation through attention.
  In: ICML (2021)

\bibitem{van2021unsupervised}
Van~Gansbeke, W., Vandenhende, S., Georgoulis, S., Van~Gool, L.: Unsupervised
  semantic segmentation by contrasting object mask proposals. In: ICCV (2021)

\bibitem{varma2019idd}
Varma, G., Subramanian, A., Namboodiri, A., Chandraker, M., Jawahar, C.: Idd: A
  dataset for exploring problems of autonomous navigation in unconstrained
  environments. In: WACV (2019)

\bibitem{wang2020deep}
Wang, J., Sun, K., Cheng, T., Jiang, B., Deng, C., Zhao, Y., Liu, D., Mu, Y.,
  Tan, M., Wang, X., et~al.: Deep high-resolution representation learning for
  visual recognition. IEEE TPAMI  (2020)

\bibitem{weston2019probably}
Weston, R., Cen, S., Newman, P., Posner, I.: Probably unknown: Deep inverse
  sensor modelling radar. In: ICRA (2019)

\bibitem{xie2021segformer}
Xie, E., Wang, W., Yu, Z., Anandkumar, A., Alvarez, J.M., Luo, P.: Segformer:
  Simple and efficient design for semantic segmentation with transformers. In:
  NeurIPS (2021)

\bibitem{xie2020propagate}
Xie, Z., Lin, Y., Zhang, Z., Cao, Y., Lin, S., Hu, H.: Propagate yourself:
  Exploring pixel-level consistency for unsupervised visual representation
  learning. In: CVPR (2021)

\bibitem{yu2020bdd100k}
Yu, F., Chen, H., Wang, X., Xian, W., Chen, Y., Liu, F., Madhavan, V., Darrell,
  T.: Bdd100k: A diverse driving dataset for heterogeneous multitask learning.
  In: CVPR (2020)

\bibitem{yuan2020object}
Yuan, Y., Chen, X., Wang, J.: Object-contextual representations for semantic
  segmentation. In: ECCV (2020)

\bibitem{zamir2018taskonomy}
Zamir, A.R., Sax, A., Shen, W., Guibas, L.J., Malik, J., Savarese, S.:
  Taskonomy: Disentangling task transfer learning. In: CVPR (2018)

\bibitem{zhang2020self}
Zhang, X., Maire, M.: Self-supervised visual representation learning from
  hierarchical grouping. In: NeurIPS (2020)

\bibitem{zhao2018sound}
Zhao, H., Gan, C., Rouditchenko, A., Vondrick, C., McDermott, J., Torralba, A.:
  The sound of pixels. In: ECCV (2018)

\bibitem{zhao2017pyramid}
Zhao, H., Shi, J., Qi, X., Wang, X., Jia, J.: Pyramid scene parsing network.
  In: CVPR (2017)

\bibitem{Zheng_2021_CVPR}
Zheng, S., Lu, J., Zhao, H., Zhu, X., Luo, Z., Wang, Y., Fu, Y., Feng, J.,
  Xiang, T., Torr, P.H., Zhang, L.: Rethinking semantic segmentation from a
  sequence-to-sequence perspective with transformers. In: CVPR (2021)

\end{thebibliography}

\clearpage
\chapter*{Appendix}
\vspace{-30pt}

\appendix
\setcounter{section}{0}
\renewcommand{\thesection}{\Alph{section}}

\renewcommand{\contentsname}{Table of Contents\vspace{-10pt}}
\addtocontents{toc}{\protect\setcounter{tocdepth}{2}}
{
\hypersetup{linkcolor=black}
\tableofcontents
}

\section{Additional quantitative results}

\subsection{Results with another training dataset}
\label{sec:nuscenes_results}

\begingroup
\begin{table}[b!]
\scriptsize
\centering
\caption{
    \textbf{Comparative results of unsupervised semantic segmentation methods when trained on nuScenes}. Comparison to the state of the art on Cityscapes~\cite{Cordts2016Cityscapes} (CS), DarkZurich~\cite{SDV20} (DZ) and Nighttime Driving~\cite{daytime:2:nighttime} (ND) datasets measured by the mean IoU (mIoU). 
    The colored differences are reported with respect to the state-of-the-art approach of~\cite{cho2021picie} denoted by 
    {\tiny \faAnchor};  
    `\emph{sup. init.}' stands for supervised initialization of the \emph{encoder} and the column
    `\emph{train. data}' indicates
    the dataset used for training,  namely nuScenes~\cite{nuscenes} (nuSC).
}
\label{tab:nuscenes_cs}
\vspace{-5pt}
\begin{tabular}{lcc|S[table-format=2.1]r|
S[table-format=2.1]r|
S[table-format=2.1]r|
S[table-format=2.1]r}
\bottomrule
 & sup. & train. &  \multicolumn{2}{c|}{CS19~\cite{Cordts2016Cityscapes}} & \multicolumn{2}{c|}{CS27~\cite{Cordts2016Cityscapes}} & \multicolumn{2}{c|}{DZ~\cite{SDV20}} & \multicolumn{2}{c}{ND~\cite{daytime:2:nighttime}} \\
\textcolor{gray}{architecture,} method & init. & data &
\multicolumn{2}{c|}{mIoU} & 
\multicolumn{2}{c|}{mIoU} & 
\multicolumn{2}{c|}{mIoU} &
\multicolumn{2}{c}{mIoU} \\
\hline\hline
\textcolor{gray}{RN18+FPN}  & & & & & & & & & & \\
{\tiny \faAnchor}\hspace{1mm}PiCIE$^\star$~\cite{cho2021picie}  & yes &  nuSC &
15.8 & &
9.7 & & 
4.6 & & 
9.9 & \\
\hspace{3mm}Modified DC$^\star$~\cite{caron2018deep} & yes & nuSC & 
11.6 &  {(\Minus4.2)} &
7.1 &  {(\Minus2.6)} & 
7.7 &  {\color{better}{(\Plus3.1)}} & 
8.3 &  {(\Minus1.6)} \\
\hspace{3mm}\ours (Ours, $\student$) & yes &  nuSC &
16.2 &  {\color{better}{(\Plus0.4)}} &
11.4 &  {\color{better}{(\Plus1.7)}} & 
7.5 &  {\color{better}{(\Plus2.9)}} &
10.2 &  {\color{better}{(\Plus0.3)}} \\

\hline
\multicolumn{3}{l|}{\textcolor{gray}{Segmenter, ViT-S/16}} & & & & & & & & \\
\hspace{3mm}\ours (Ours, $\student$) & no &  nuSC &
19.8 &  {\color{better}{(\Plus4.0})} &
13.9 &  {\color{better}{(\Plus4.2})} &
9.7 &  {\color{better}{(\Plus5.1})} &
14.1 &  {\color{better}{(\Plus4.2})} \\
\toprule
\multicolumn{11}{l}{\textsuperscript{$*$} Our training using PiCIE code base.}
\end{tabular}
\end{table}
\endgroup

\begingroup
\setlength{\tabcolsep}{0.6pt} 
\begin{table}[ht!]
\centering
\caption{\textbf{Comparative results on ACDC when methods trained on nuScenes}. Comparison to the state of the art %
for unsupervised semantic segmentation on the ACDC~\cite{SDV21} dataset. Please refer to Table~\ref{tab:nuscenes_cs} for the symbols.
}
\vspace{-5pt}
\label{tab:nuscenes_acdc}
\scriptsize
\begin{tabular}{lcc|
S[table-format=2.1]r|
S[table-format=2.1]r|
S[table-format=2.1]r|
S[table-format=2.1]r||
S[table-format=2.1]r}
\bottomrule
& sup. & train. & \multicolumn{2}{c|}{night} & \multicolumn{2}{c|}{fog} & \multicolumn{2}{c|}{rain} & \multicolumn{2}{c||}{snow}& \multicolumn{2}{c}{average} \\
\textcolor{gray}{architecture,} method & init. & data &
\multicolumn{2}{c|}{mIoU} &
\multicolumn{2}{c|}{mIoU} &
\multicolumn{2}{c|}{mIoU} &
\multicolumn{2}{c||}{mIoU} &
\multicolumn{2}{c}{mIoU} \\
\hline\hline
\textcolor{gray}{RN18+FPN} & & & & & & & & & & & & \\
{\tiny \faAnchor}\hspace{1mm}PiCIE$^*$~\cite{cho2021picie} & yes & nuSC &
4.3 & \scriptsize{\color{better}{}} & 
8.9 & \scriptsize{\color{better}{}} & 
9.5 & \scriptsize{\color{better}{}} & 
7.5 & \scriptsize{\color{better}{}} & 
7.5 & \scriptsize{\color{better}{}}
\\
\hspace{2.5mm}Modified DC$^*$~\cite{caron2018deep} & yes & nuSC &
6.7 & \scriptsize{\color{better}{(\Plus2.4})} & 
11.7 & \scriptsize{\color{better}{(\Plus2.8)}} & 
10.4 & \scriptsize{\color{better}{(\Plus0.9)}} & 
9.6 & \scriptsize{\color{better}{(\Plus2.1)}} & 
9.6 & \scriptsize{\color{better}{(\Plus2.1)}}
\\
\hspace{2.5mm}\ours (Ours, $\student$) & yes & nuSC &
7.9 & \scriptsize{\color{better}{(\Plus3.6)}} & 
14.3 & \scriptsize{\color{better}{(\Plus5.4)}} & 
14.4 & \scriptsize{\color{better}{(\Plus4.9)}} & 
13.4 & \scriptsize{\color{better}{(\Plus5.9)}} & 
12.5 & \scriptsize{\color{better}{(\Plus5.0)}}
\\
\hline

\multicolumn{3}{l|}{\textcolor{gray}{Segmenter, ViT-S/16}} & & & & &  & & & & & \\
\hspace{2.5mm}\ours (Ours, $\student$) & no & nuSC &
10.6 & \scriptsize{\color{better}{(\Plus6.3)}} & 
13.3 & \scriptsize{\color{better}{(\Plus4.4)}} & 
16.0 & \scriptsize{\color{better}{(\Plus6.5)}} & 
14.8 & \scriptsize{\color{better}{(\Plus7.3)}} & 
13.9 & \scriptsize{\color{better}{(\Plus6.4)}}
\\
\toprule
\end{tabular}
\end{table}
\endgroup

We report in Tables~\ref{tab:nuscenes_cs} and~\ref{tab:nuscenes_acdc} %
the performance of \ours when trained 
using a subset of ${\sim}8$k images from the nuScenes dataset \cite{nuscenes}. 
As shown in Table~\ref{tab:nuscenes_cs}, 
the mIoU on Cityscapes is $19.8$.
Although there is a small drop from the $21.8$ achieved with \ours trained on Waymo Open, the results are still significantly better than those of the competing methods.
This drop might be caused by differences of statistics between the two datasets, \eg, nuScenes has fewer examples of smaller-object classes, such as pedestrians.

\subsection{Ablation of the number of clusters in unsupervised labeling}
\label{sec:clusters_number_ablation}

\begin{figure}[t]
        \centering
        \begin{tikzpicture}
    \begin{axis}[%
        axis x line=bottom,
        axis y line=left,
        xlabel= $\#$ clusters,
        ylabel=mIoU,
        width=5.6cm,
        height=4.1cm,
        xtick={20,25,30,35,40},
        xmax=43,
        ymax=20.8,
        x label style={at={(axis description cs:0.4,-0.15)},anchor=north}, 
        y label style={at={(axis description cs:-0.12,.5)},anchor=south},
        label style={font=\footnotesize},
        tick label style={font=\footnotesize},  
        legend style={font=\footnotesize,at={(1,1)},anchor=north west,draw=none}  
        ]
        \addplot[mark=none,RoyalPurple,ultra thick] coordinates {(20, 20.0) (25, 20.1) (30, 20.4) (35, 19.6) (40, 17.0)}; %
        \addlegendentry{Segmenter}
        \addplot[mark=none,BrickRed,ultra thick] coordinates {(20, 16.5) (25, 16.4) (30, 17.4) (35, 16.5) (40, 15.7)}; %
        \addlegendentry{RN18\Plus FPN}
    \end{axis}
\end{tikzpicture}
    \caption{        
    \textbf{Ablation of the number of clusters}. Performance in mIoU, when using the
        {\textcolor{RoyalPurple}{\textbf{Segmenter}}} model and the {\textcolor{BrickRed}{\textbf{ResNet18\Plus FPN}}} model on the Cityscapes dataset,  
        as a function of the number of clusters in the unsupervised labeling step.  
        }
    \label{fig:num_clusters_comparison}
\end{figure}
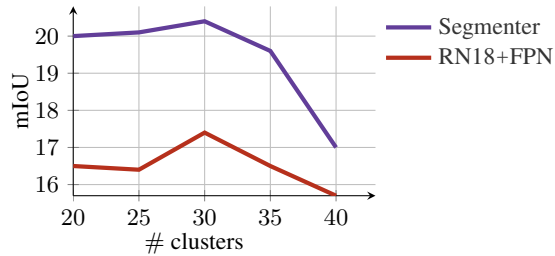

Here we investigate the sensitivity of our method to the number $k$ of clusters used for unsupervised labeling.
Figure~\ref{fig:num_clusters_comparison} shows the mIoU results on Cityscapes for~$k\,{\in}\,\{20,25,30,35,40\}$. 
In all cases we use a ViT-S/16 feature extractor trained with DINO.
The results show that for $k\,{\in}\,\{20,25,30,35\}$ the \emph{mIoU performance is fairly stable}.
As expected, when the number of clusters becomes much higher than the number of Cityscapes classes (\eg, $k=40$), the performance drops.

\subsection{Influence of the LiDAR's density}
\label{sec:lidar_density_ablation}
We investigate here the performance of \ours %
when provided with sparser LiDAR data. We performed experiments on the Waymo Open dataset and downsampled the LiDAR data from $64$ to $32$ beam channels by dropping every other channel. We re-trained the \emph{teacher} model three times and report the average performance (following the setup of the main paper). 
We obtained $20.3$ mIoU, which is only slightly lower than the $20.4$ obtained with the full LiDAR resolution, demonstrating the robustness of our method to this considerable decrease of LiDAR resolution.
However, as already discussed in the main paper, our method will likely not work well with extremely sparse LiDAR data (\eg, low-cost LiDARs with 4-beam channels). 
Such a sparsity would lead to poor LiDAR-based segments and geometric priors that would rather confuse the model, instead of teaching it to recognize objects.

\subsection{Pixel accuracy results}
\label{sec:pixel_accuracy}
In Tables~\ref{tab:pa_cs} and~\ref{tab:pa_acdc}, we report results measured with the pixel accuracy (PA) metric corresponding to all experiments of our main paper. 
We observe that results follow a similar trend to those measured with mIoU.
\begingroup
\setlength{\tabcolsep}{1.0pt} 
\begin{table*}[t!]
\centering
\caption{
    \textbf{Comparative results using PA metric}. Comparison to the state of the art
    for unsupervised semantic segmentation on Cityscapes~\cite{Cordts2016Cityscapes} (CS), DarkZurich~\cite{SDV20} (DZ) and Nighttime driving~\cite{daytime:2:nighttime} (ND) datasets measured by the pixel accuracy (PA). 
    Same organization as Table \ref{tab:nuscenes_cs}. For easy reference, rows are colored according to the used training dataset. 
}
\label{tab:pa_cs}
\scriptsize
\begin{tabular}{lcc|
S[table-format=2.1]r|S[table-format=2.1]r|
S[table-format=2.1]r|S[table-format=2.1]r
}
\bottomrule
 & sup. & train. &  \multicolumn{2}{c|}{CS19~\cite{Cordts2016Cityscapes}} & \multicolumn{2}{c|}{CS27~\cite{Cordts2016Cityscapes}} & \multicolumn{2}{c|}{DZ~\cite{SDV20}} & \multicolumn{2}{c}{ND~\cite{daytime:2:nighttime}} \\
\textcolor{gray}{architecture,} method & init. & data &
\multicolumn{2}{c|}{PA} & 
\multicolumn{2}{c|}{PA} & 
\multicolumn{2}{c|}{PA} &
\multicolumn{2}{c}{PA} \\
\hline\hline
\textcolor{gray}{RN18+FPN} & & & & & & & & & & \\
\rowcolor{Gray!10!white}{\tiny \faAnchor}\hspace{1mm}PiCIE$^\ddagger$~\cite{cho2021picie} & yes &  CS &
63.1 & & 62.7 & & 30.7 & & 41.4 & \\
\rowcolor{Gray!10!white}\hspace{3mm}IIC$^\dagger$~\cite{ji2019invariant} & yes & CS &
\multicolumn{2}{c|}{\text{-}} & 
47.9 & {(\Minus14.8)} & \multicolumn{2}{c|}{\text{-}} & \multicolumn{2}{c}{\text{-}}  \\
\rowcolor{Gray!10!white}\hspace{3mm}Modified DC$^\ddagger$~\cite{caron2018deep} & yes &  CS & 
52.4 & {(\Minus10.7)} & 
52.1 & {(\Minus10.7)} & 
42.4 & { \color{better}{(\Plus11.7)}} &
46.2 & { \color{better}{(\Plus4.8)}}
\\

\hspace{3mm}Modified DC$^\star$ & yes & nuSc & 
45.9 & {(\Minus17.2)} &
45.7 & {(\Minus17.0)} &
41.4 & { \color{better}{(\Plus10.7)}} &
41.9 & { \color{better}{(\Plus0.5)}} \\
\hspace{3mm}PiCIE$^\star$ & yes &  nuSc &
61.6 & {(\Minus1.5)} &
61.3 & {(\Minus1.4)} &
29.6 & {(\Minus1.1)} &
45.1 & { \color{better}{(\Plus3.7)}} \\
\hspace{3mm}\ours (Ours, $\student$) & yes &  nuSc &
61.4 & {(\Minus1.7)} &
61.1 & {(\Minus1.6)} & 
37.4 & { \color{better}{(\Plus6.7)}} & 
33.6 & {(\Minus7.8)} \\

\rowcolor{Green!6!white}\hspace{3mm}Modified DC$^\star$ & yes & WO & 
55.6 & {(\Minus7.5)} & 
43.2 & {(\Minus19.5)} & 
35.8 &  \color{better}{(\Plus5.1)} &
33.4 & {(\Minus8.0)} \\
\rowcolor{Green!6!white}\hspace{3mm}PiCIE$^\star$ & yes & WO & 
48.6 & {(\Minus14.5)} & 
48.3 & {(\Minus14.4)} & 
31.9 & { \color{better}{(\Plus1.1)}} & 
40.0 & {(\Minus1.4)} \\
\rowcolor{Green!6!white}\hspace{3mm}\ours (Ours, $\student$) 
& yes &  WO &
66.4 & { \color{better}{(\Plus3.3)}} &
67.1 & { \color{better}{(\Plus4.3)}} & 
47.7 & { \color{better}{(\Plus17.0)}} & 
49.0 & { \color{better}{(\Plus7.6)}} \\

\hline
\multicolumn{3}{l|}{\textcolor{gray}{Segmenter, ViT-S/16}} & & & & & & & & \\
\hspace{3mm}\ours (Ours, $\student$) & no &  nuSc &
73.2 & {\color{better}{(\textbf{\Plus10.1})}} &
72.8 & {\color{better}{(\textbf{\Plus10.1})}} &
50.2 & {\color{better}{(\Plus19.5)}} &
65.5 & {\color{better}{(\textbf{\Plus24.1})}} \\
\rowcolor{Green!6!white}\hspace{3mm}\ours (Ours, $\student$) 
& no & WO &
\textbf{69.5} & { \color{better}{(\Plus6.4})} &
\textbf{69.1} & { \color{better}{(\Plus6.4)}} &
\textbf{55.9} & { \color{better}{(\Plus25.1})} &
\textbf{60.2} & { \color{better}{(\Plus18.8)}} \\
\toprule
\multicolumn{11}{l}{\scriptsize \textsuperscript{$\dagger$} Results reported in~\cite{cho2021picie}. \textsuperscript{$\ddagger$} Models provided by the PiCIE~\cite{cho2021picie} authors.} \\
\multicolumn{11}{l}{\textsuperscript{$*$} Trained by PiCIE code base.}
\end{tabular}
\end{table*}
\endgroup
\begingroup
\setlength{\tabcolsep}{0.4pt} 
\begin{table}[t!]
\centering
\caption{
\textbf{Comparative results on ACDC using PA metric}. 
Comparison to the state-of-the-art approach~\cite{cho2021picie} for unsupervised semantic segmentation on the ACDC~\cite{SDV21} dataset. 
Same organization as Table \ref{tab:nuscenes_acdc}. For easy reference, rows are colored according to the used training dataset
}
\label{tab:pa_acdc}
\scriptsize
\begin{tabular}{lcc|
S[table-format=2.1]r|S[table-format=2.1]r|
S[table-format=2.1]r|S[table-format=2.1]r||
S[table-format=2.1]r}
\bottomrule
& sup. & train. & \multicolumn{2}{c|}{night} & \multicolumn{2}{c|}{fog} & \multicolumn{2}{c|}{rain} & \multicolumn{2}{c||}{snow} & \multicolumn{2}{c}{average} \\
method & init. & data &
\multicolumn{2}{c|}{PA} & 
\multicolumn{2}{c|}{PA} & 
\multicolumn{2}{c|}{PA} &
\multicolumn{2}{c||}{PA} &
\multicolumn{2}{c}{PA} \\
\hline\hline
\textcolor{gray}{RN18+FPN} & & & & & & & & & & & & \\

\rowcolor{Gray!10!white}{\tiny \faAnchor}\hspace{1mm}PiCIE~\cite{cho2021picie} & yes & CS & 
25.8 & & 50.0 & & 53.6 & & 50.4 & & 45.0 &  \\
\rowcolor{Gray!10!white}\hspace{3mm}MDC~\cite{caron2018deep} & yes & CS &
43.0 & {\color{better}{(\Plus17.3)}} & 
43.6 & {(\Minus6.4)} & 
35.0 & {(\Minus18.6)} & 
38.8 & {(\Minus11.5)} & 
40.1 & {(\Minus4.8)} \\

\hspace{3mm}Modified DC$^*$ & yes & nuSC &
36.5 & {\color{better}{(\Plus10.7)}} & 
44.8 & {(\Minus5.2)} & 
41.4 & {(\Minus12.2)} &
38.5 & {(\Minus11.9)} &
40.3 & {(\Minus4.7)}
\\
\hspace{3mm}PiCIE$^*$ & yes & nuSC &
26.9 & {\color{better}{{(\Plus1.1)}}} &
33.1 & {(\Minus16.9)} &
33.4 & {(\Minus20.2)} &
29.1 & {(\Minus21.3)} &
30.6 & {(\Minus14.4)}
\\
\hspace{3mm}\ours (Ours, $\student$) & yes & nuSC &
34.5 & {\color{better}{{(\Plus8.7)}}} & 
59.4 & {\color{better}{{(\Plus9.4)}}} & 
58.2 & {\color{better}{{(\Plus4.6)}}} & 
53.9 & {\color{better}{{(\Plus3.5)}}} & 
51.5 & {\color{better}{{(\Plus6.5)}}}
\\

\rowcolor{Green!6!white}\hspace{3mm}MDC$^\star$ & yes & WO &
32.9 & \color{better}{{(\Plus7.2)}} & 
47.0 & {(\Minus3.0)} & 
40.3 & {(\Minus13.3)} & 
44.2 & {(\Minus6.2)} & 
41.1 & {(\Minus3.8)} \\
\rowcolor{Green!6!white}\hspace{3mm}PiCIE$^\star$ & yes & WO &
27.2 & \color{better}{{(\Plus1.4)}} & 
{56.9} & \color{better}{{{(\Plus6.8)}}} & 
53.8 & \color{better}{{(\Plus0.2)}} & 
53.0 & \color{better}{{(\Plus2.6)}} & 
47.7 & \color{better}{{(\Plus2.8)}} \\
\rowcolor{Green!6!white}\hspace{3mm}\ours (Ours, $\student$) 
& yes & WO &
43.2 & {\color{better}{(\Plus17.5)}} &
56.5 & {\color{better}{(\Plus6.5)}} & 
{54.1} & {\color{better}{{(\Plus0.5)}}} & 
55.5 & {\color{better}{(\Plus5.1)}} & 
52.3 & {\color{better}{(\Plus7.4)}} \\

\hline
\multicolumn{3}{l|}{\textcolor{gray}{Segmenter, ViT-S/16}} & & & & & & & & & & \\
\hspace{3mm}\ours (Ours, $\student$) & no & nuSC &
50.2 & \color{better}{(\Plus24.4)} & 
60.2 & \color{better}{(\Plus10.2)} & 
62.5 & \color{better}{(\Plus8.9)} & 
56.5 & \color{better}{(\Plus6.1)} & 
57.5 & \color{better}{(\Plus12.5)}
\\

\rowcolor{Green!6!white}\hspace{3mm}\ours (Ours, $\student$) 
& no & WO &
\bf{52.6} & {\color{better}{{(\Plus26.9)}}} &
54.2 & {\color{better}{(\Plus4.2)}} & 
50.1 & {(\Minus3.5)} & 
\bf{56.8} & {\color{better}{{(\Plus6.4)}}}& 
\bf{53.4} & {\color{better}{{(\Plus8.5)}}} \\
\hline
\end{tabular}
\end{table}
\endgroup

\subsection{Category-wise results}
\label{sec:per_class_results}
\begin{table*}[t!]
\caption{
\textbf{Per-class comparative performance on Cityscapes}. Per-class IoU is evaluated using the Hungarian algorithm on the $19$ validation classes. We can see significant benefits of \ours (`\das') over PiCIE in $14$ (including all road users and objects) out of $19$ classes. \ours works much worse for \emph{sidewalk} and \emph{sky} as we discuss in Sections~\ref{sec:per_class_results} and~\ref{sec:failure_cases}. `\texttt{(CS)}' stands for a model trained on the Cityscapes~\cite{Cordts2016Cityscapes} dataset, while `\texttt{(WO)}' for models trained on the Waymo Open~\cite{sun2020scalability} dataset. The best results per class are highlighted in bold and color.
}
\resizebox{1.0\columnwidth}{!}{
\scriptsize
\centering
\begin{tabular}{l|r|r|r|r|r|r|r|r|r|r|r|r|r|r|r|r|r|r|r||r}
\bottomrule
& \multicolumn{1}{c|}{\rotatebox[origin=l]{90}{road}} &
\multicolumn{1}{c|}{\rotatebox[origin=l]{90}{sidewalk}} &
\multicolumn{1}{c|}{\rotatebox[origin=l]{90}{building}} &
\multicolumn{1}{c|}{\rotatebox[origin=l]{90}{wall}} &
\multicolumn{1}{c|}{\rotatebox[origin=l]{90}{fence}} &
\multicolumn{1}{c|}{\rotatebox[origin=l]{90}{pole}} &
\multicolumn{1}{c|}{\rotatebox[origin=l]{90}{traffic light}} &
\multicolumn{1}{c|}{\rotatebox[origin=l]{90}{traffic sign}} &
\multicolumn{1}{c|}{\rotatebox[origin=l]{90}{vegetation}} &
\multicolumn{1}{c|}{\rotatebox[origin=l]{90}{terrain}} &
\multicolumn{1}{c|}{\rotatebox[origin=l]{90}{sky}} &
\multicolumn{1}{c|}{\rotatebox[origin=l]{90}{person}} &
\multicolumn{1}{c|}{\rotatebox[origin=l]{90}{rider}} &
\multicolumn{1}{c|}{\rotatebox[origin=l]{90}{car}} &
\multicolumn{1}{c|}{\rotatebox[origin=l]{90}{truck}} &
\multicolumn{1}{c|}{\rotatebox[origin=l]{90}{bus}} &
\multicolumn{1}{c|}{\rotatebox[origin=l]{90}{train}} &
\multicolumn{1}{c|}{\rotatebox[origin=l]{90}{~motorcycle~}} &
\multicolumn{1}{c||}{\rotatebox[origin=l]{90}{bicycle}}& 
\multicolumn{1}{c}{\rotatebox[origin=l]{90}{mIoU}} \\
\hline
\textcolor{gray}{RN18+FPN} &&&&&&&& & & & & & & & & & & & &  \\
PiCIE~\cite{cho2021picie} (\texttt{CS}) & 58.2 & 12.5 & 63.8 & ~1.0 & ~2.4 & ~1.3 & ~0.1 & ~0.4 & 55.5 & ~1.7 & 44.7 & ~1.9 & ~0.5 & 48.2 & ~1.3 & ~3.9 & ~1.0 & ~0.5 & ~1.6 & 15.8 \\

PiCIE~\cite{cho2021picie} \texttt{(WO)} & 58.5 & \cellcolor{bestbg}\textbf{13.8} & 35.8 & \cellcolor{bestbg}\textbf{6.7} & 0.7 & 1.2 & 0.4 & 1.2 & 28.3 & 1.2 & \cellcolor{bestbg}\textbf{55.8} & 3.1 & 0.6 & 48.5 & 0.5 & 1.5 & 0.3 & 0.0 & 2.3 & 13.7 \\
\das (Ours, \texttt{WO}) & 72.7 & 7.0 & 56.6 & 4.5 & \cellcolor{bestbg}\textbf{5.6} & 16.9 & 3.6 & 15.7 & \cellcolor{bestbg}\textbf{66.8} & \cellcolor{bestbg}\textbf{2.2} & 6.0 & 40.0 & \cellcolor{bestbg}\textbf{5.0} & 44.7 & 0.5 & 18.5 & 0.2 & \cellcolor{bestbg}\textbf{1.4} & 2.1 & 19.5 \\
\hline
\multicolumn{2}{l|}{\textcolor{gray}{Segmenter, ViT-S/16}} &&&&&&& & & & & & & & & & & & &  \\
\das (Ours, \texttt{WO}) & \cellcolor{bestbg}\textbf{74.1} & 7.0 & \cellcolor{bestbg}\textbf{65.7} & \cellcolor{bestbg}\textbf{6.6} & 1.0 & \cellcolor{bestbg}\textbf{24.9} & \cellcolor{bestbg}\textbf{4.3} & \cellcolor{bestbg}\textbf{16.6} & 64.8 & 1.8 & 3.7 & \cellcolor{bestbg}\textbf{45.9} & 4.3 & \cellcolor{bestbg}\textbf{57.3} & \cellcolor{bestbg}\textbf{1.7} & \cellcolor{bestbg}\textbf{19.9} & \cellcolor{bestbg}\textbf{1.3} & 0.4 & \cellcolor{bestbg}\textbf{12.1} & \cellcolor{bestbg}\textbf{21.8} \\
\toprule
\end{tabular}
}
\label{table:cityscapes_class_iou_v2}
\end{table*}

In the main paper, we have presented results averaged over all classes. We report in Table~\ref{table:cityscapes_class_iou_v2} the \emph{per-class IoU} results of our \ours approach on the Cityscapes dataset. 

We observe that \ours outperforms the baseline PiCIE on 15 out of 19 classes. IoU gains (w.r.t. PiCIE trained on Waymo Open dataset) are significant for small-object classes such as \emph{pole} ($\Plus 23.2/{\Plus }15.2$ with Segmenter and ResNet18\Plus FPN respectively), \emph{traffic signs} ($\Plus 15.4/{\Plus }14.5$), and \emph{person} ($\Plus 42.8/{\Plus }36.9$). They are also  substantial for some classes that can cover larger image portions, \eg, \emph{road} ($\Plus 15.6/{\Plus }14.2$), \emph{vegetation} ($\Plus 36.5/{\Plus }38.5$), \emph{car} ($\Plus 8.8/{\Minus}3.8$).
The results of ResNet18\Plus FPN are slightly worse on the \emph{car} class because 
\emph{car} instances %
are split into several pseudo-classes.
Gains over \emph{road} and \emph{car} were expected since LiDAR data provide very good segments for these classes; 
it is more surprising to see gains on \emph{vegetation}, a class that is not easily captured by LiDAR.

\section{Analyzing learned representations}

\subsection{$k$-NN evaluation of learned representations}
\label{sec:knn_eval}
To evaluate the quality of the learned representations, we compare the representations produced by a ResNet18 backbone trained (a) on Imagenet in a fully-supervised fashion for the classification task, 
(b) using PiCIE~\cite{cho2021picie} trained on Waymo Open, and (c) using our \ours trained on Waymo Open.
For this comparison we perform $k$-NN based pixel-wise classification on the Cityscapes validation set using a \emph{low-shot scenario} where only $100$ Cityscapes training images are available (we consider three random splits of 100 images from~\cite{french2020semi} and report the average results). 
Our goal is to analyze the ability of the representations to learn with few training examples.
In Table~\ref{tab:knn_eval_wDiff}, we report results in terms of pixel accuracy 
for $k\,{\in}\,\{1,5,20\}$ and observe that \ours outperforms both the supervised baseline and PiCIE~\cite{cho2021picie}.

\begin{table}[t]
    \centering
    \caption{\textbf{Evaluation of learned features using $k$-NN pixel-wise classification}. %
    Results are produced by running $k$-NN with three different $100$-image training sets \cite{french2020semi} and computing the average (over the three runs) pixel accuracy on the Cityscapes validation split. Results are reported with the Pixel Accuracy (PA) metric.
    }
    \label{tab:knn_eval_wDiff}
    \begin{tabular}{l|cr|cr|cr}
    \toprule
    method & \multicolumn{2}{c|}{$k=1$} & \multicolumn{2}{c|}{$k=5$} & \multicolumn{2}{c}{$k=20$} \\
    \hline
    supervised & 76.9 & & 79.4 & & 81.2 & \\
    PiCIE~\cite{cho2021picie} & 74.3 & (-2.6) & 78.0 & (-1.4) & 79.1 & (-2.1) \\
    \ours & 81.1 & {\color{better}{(+4.2)}} & 83.2 & {\color{better}{(+3.8)}} & 84.7 & {\color{better}{(+3.5)}} \\
    \bottomrule
    \end{tabular}
\end{table}

\subsection{Representation analysis via PCA}

\begin{figure*}[t!]
    \centering
    \includegraphics[width=0.25\linewidth]{img/qualitative_new/frankfurt_000001_011835_leftImg8bit_orig.jpg}%
    \includegraphics[width=0.25\linewidth]{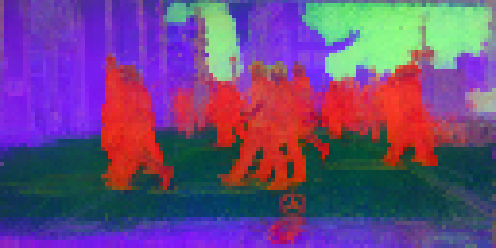}%
    \includegraphics[width=0.25\linewidth]{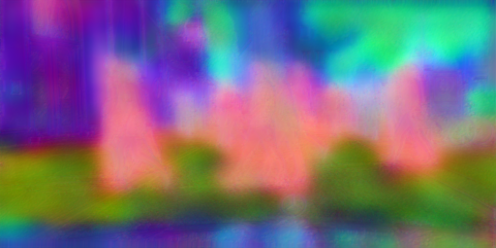}%
    \includegraphics[width=0.25\linewidth]{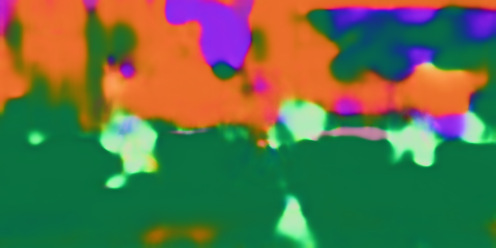}
    \includegraphics[width=0.25\linewidth]{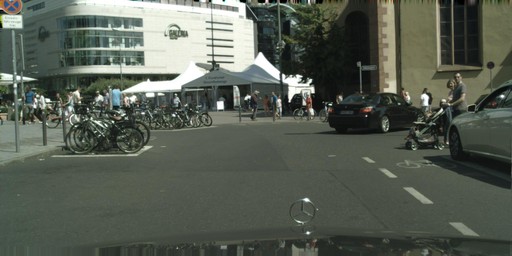}%
    \includegraphics[width=0.25\linewidth]{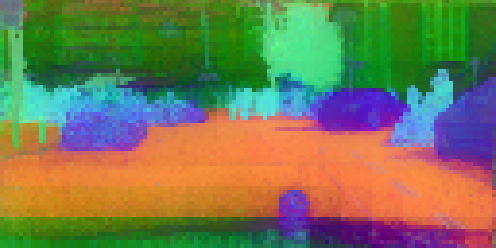}%
    \includegraphics[width=0.25\linewidth]{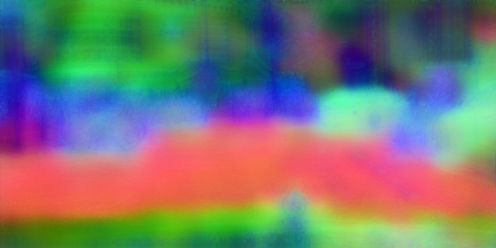}%
    \includegraphics[width=0.25\linewidth]{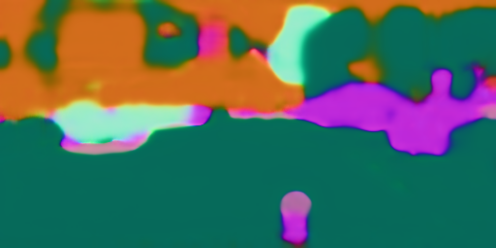}
    {
    \hspace*{0.6cm}
    {input}\hspace{1.6cm}
    {ours (Segm.)}\hspace{1.35cm}
    {ours (RN)}\hspace{1.35cm}
    {PiCIE (RN)}
    }
    \caption{\textbf{Feature visualization}. 
    We do PCA analysis of the pixel-wise decoder features from each image (independenly between the different images) and visualize the three first PCA components as an RGB image.
    `Segm.' stands for Segmenter with ViT-S/16 and `RN' for ResNet18+FPN.}
    \label{fig:pca_features}
\end{figure*}

In Figure~\ref{fig:pca_features}, we visualize the three main PCA components of the \emph{decoder} features as RGB. We observe that our features learned with Segmenter separate better object classes.

\subsection{Confusion matrices for class mapping}
\label{sec:confusion}
Here we analyze the confusion matrices, presented in Figure~\ref{fig:confusion_matrices_row_reordered1}, %
which provide the mapping between ground truth and pseudo classes.
For each confusion matrix, we reorder the columns based on the matching obtained from the Hungarian algorithm, and $L_1$-normalize the values per row, i.e., per ground-truth class (for simplicity, we do not illustrate the un-matched pseudo-classes in the figures).
Thus, a value of $1$ would signify that all pixels in a ground-truth class belong to a single pseudo-class. 
Moreover, due to the reordering, the largest values should ideally be on the diagonal of the confusion matrix. %

For each row, the highest and the diagonal entry are reported. We note that, for \ours (Fig.~\ref{fig:confusion_matrices_row_reordered1}(a)), 90\% of the road pixels are covered by the first pseudo-class. However, this pseudo-class also covers large portions of sidewalk and vegetation as all these labels belong to ground pixels and hence are segmented together by our LiDAR-based segment proposal mechanism. Similarly, pseudo-class 12 overlaps person, rider, motorcycle and bicycle, i.e., with human-related ground-truth classes. Regarding PiCIE (Fig.~\ref{fig:confusion_matrices_row_reordered1}(b)), only a few pseudo-classes 
have a significant overlap with ground-truth classes. %
In particular, the pseudo-class 3
overlaps with the majority of the ground-truth classes. 

\setlength{\fboxsep}{3mm} %
\renewcommand{\tabcolsep}{-2.pt}
\begin{figure*}[t!]
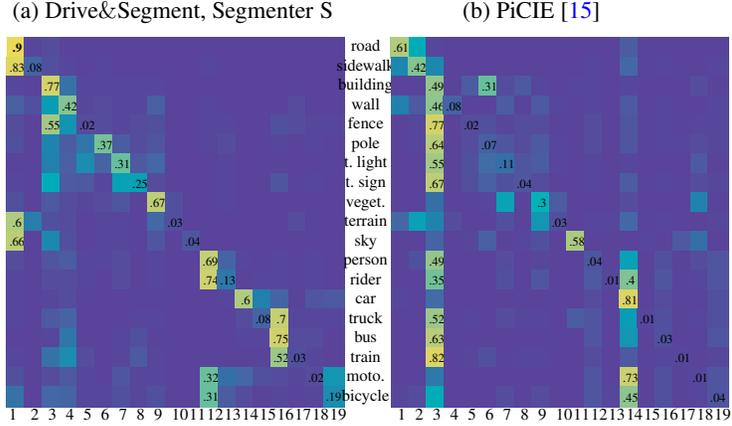

    \centering
    \subfloat[][\ours, Segmenter~$\student$\label{fig:confusion_segmenter_row_reordered}]{
    \resizebox{!}{.216\linewidth}{
    \large
      \begin{tabular}{l*{19}{G}}
    \vspace{-.49em}
    \pgfmathsetmacro{\PercentColor}{max(min(100.0*(.9 - \MinNumber)/(\MaxNumber-\MinNumber),100.0),0.00)} %
    \hspace{-0.65em}
    \setlength{\fboxsep}{-.1pt}
    \colorbox{Goldenrod!\PercentColor!BlueGreen}{\makebox(16,16){\centering \footnotesize \textbf{\stripzero{.9}}}}
    & .01 & .02 & .0 & .0 & .0 & .0 & .0 & .0 & .0 & .0 & .0 & .0 & .0 & .0 & .0 & .0 & .0 & .0  \\ 
    \vspace{-.49em}
    \ApplyGradient{.83} & -.08 & .03 & .01 & .0 & .01 & .0 & .0 & .0 & .0 & .0 & .01 & .0 & .0 & .0 & .0 & .0 & .0 & .0 \\
    \vspace{-.49em}
    \ApplyGradient{.01} & .0 & -.77 & .12 & .01 & .01 & .01 & .01 & .01 & .0 & .0 & .0 & .0 & .0 & .0 & .0 & .0 & .0 & .0 \\
    \vspace{-.49em}
    \ApplyGradient{.07} & .03 & .24 & -.42 & .01 & .02 & .01 & .01 & .07 & .0 & .0 & .01 & .0 & .0 & .0 & .0 & .02 & .0 & .0  \\
    \vspace{-.49em}
    \ApplyGradient{.02} & .0 & .55 & .21 & -.02 & .02 & .01 & .02 & .03 & .0 & .0 & .01 & .01 & .0 & .0 & .03 & .02 & .01 & .01 \\
    \vspace{-.49em}
    \ApplyGradient{.03} & .0 & .17 & .06 & .13 & -.37 & .06 & .02 & .04 & .0 & .0 & .01 & .02 & .0 & .0 & .01 & .0 & .01 & .0 \\
    \vspace{-.49em}
    \ApplyGradient{.0} & .0 & .16 & .07 & .19 & .08 & -.31 & .05 & .09 & .0 & .0 & .0 & .01 & .0 & .0 & .0 & .0 & .0 & .0 \\
    \vspace{-.49em}
    \ApplyGradient{.01} & .0 & .28 & .06 & .05 & .01 & .25 & -.25 & .02 & .0 & .0 & .0 & .01 & .0 & .0 & .04 & .0 & .01 & .0 \\
    \vspace{-.49em}
    \ApplyGradient{.03} & .0 & .05 & .02 & .01 & .02 & .04 & .0 & -.67 & .03 & .0 & .0 & .01 & .0 & .0 & .0 & .0 & .0 & .0 \\
    \vspace{-.49em}
    \ApplyGradient{.6} & .15 & .03 & .02 & .0 & .01 & .01 & .0 & .09 & -.03 & .0 & .0 & .0 & .0 & .0 & .0 & .02 & .0 & .0 \\
    \vspace{-.49em}
    \ApplyGradient{.66} & .0 & .18 & .03 & .0 & .01 & .01 & .0 & .04 & .0 & -.04 & .0 & .0 & .0 & .0 & .0 & .0 & .0 & .0   \\
    \vspace{-.49em}
    \ApplyGradient{.02} & .0 & .04 & .06 & .01 & .01 & .02 & .0 & .01 & .0 & .0 & -.69 & .08 & .01 & .0 & .01 & .0 & .01 & .0 \\
    \vspace{-.49em}
    \ApplyGradient{.01} & .0 & .02 & .02 & .01 & .0 & .02 & .0 & .01 & .0 & .0 & .74 & -.13 & .0 & .01 & .01 & .0 & .01 & .0 \\
    \vspace{-.49em}
    \ApplyGradient{.01} & .0 & .01 & .02 & .0 & .0 & .0 & .0 & .0 & .0 & .0 & .0 & .01 & -.6 & .16 & .08 & .0 & .04 & .05 \\
    \vspace{-.49em}
    \ApplyGradient{.01} & .0 & .05 & .08 & .0 & .0 & .0 & .01 & .01 & .0 & .0 & .0 & .0 & .02 & -.08 & .7 & .01 & .01 & .01 \\
    \vspace{-.49em}
    \ApplyGradient{.0} & .0 & .01 & .13 & .0 & .0 & .01 & .03 & .0 & .0 & .0 & .0 & .0 & .01 & .01 & -.75 & .02 & .02 & .0 \\
    \vspace{-.49em}
    \ApplyGradient{.01} & .0 & .12 & .19 & .02 & .0 & .01 & .05 & .01 & .0 & .0 & .0 & .0 & .0 & .0 & .52 & -.03 & .01 & .0 \\
    \vspace{-.49em}
    \ApplyGradient{.03} & .0 & .02 & .05 & .01 & .0 & .04 & .01 & .01 & .0 & .0 & .32 & .11 & .09 & .03 & .02 & .01 & -.02 & .22 \\
    \vspace{-.49em}
    \ApplyGradient{.12} & .0 & .09 & .08 & .01 & .01 & .03 & .01 & .03 & .0 & .0 & .31 & .06 & .02 & .0 & .01 & .0 & .01 & -.19 \\
          
    \rotz{1} & \rotz{2} &  \rotz{3} & \rotz{4} &  \rotz{5} & \rotz{6} & \rotz{7} & \rotz{8} & \rotz{9} &  \rotz{10} & \rotz{11} & \rotz{12} & \rotz{13} & \rotz{14} & \rotz{15} & \rotz{16} & \rotz{17} & \rotz{18} & \rotz{19} \\
    \end{tabular}
    }
} 
\hspace{-1.09em}%
\subfloat[][PiCIE~\cite{cho2021picie}]{
    \resizebox{!}{.216\linewidth}{
    \large
      \begin{tabular}{c*{20}{G}}
    \vspace{-0.49em}
    \raisebox{+.4\normalbaselineskip}[0pt][0pt]{road} & 
    -0.61 & 0.28 & 0.01 & 0.0 & 0.0 & 0.0 & 0.0 & 0.0 & 0.0 & 0.0 & 0.0 & 0.0 & 0.0 & 0.06 & 0.0 & 0.0 & 0.0 & 0.0 & 0.0 \\
    \vspace{-0.49em}
    \raisebox{+.4\normalbaselineskip}[0pt][0pt]{sidewalk} & 
    0.18 & -0.42 & 0.18 & 0.0 & 0.0 & 0.0 & 0.0 & 0.0 & 0.0 & 0.0 & 0.0 & 0.01 & 0.0 & 0.09 & 0.0 & 0.01 & 0.0 & 0.0 & 0.01 \\
    \vspace{-0.49em}
    \raisebox{+.4\normalbaselineskip}[0pt][0pt]{building} & 
    0.0 & 0.0 & -0.49 & 0.0 & 0.06 & 0.31 & 0.02 & 0.05 & 0.01 & 0.0 & 0.01 & 0.01 & 0.0 & 0.01 & 0.0 & 0.0 & 0.0 & 0.02 & 0.0 \\
    \vspace{-0.49em}
    \raisebox{+.4\normalbaselineskip}[0pt][0pt]{wall} & 
    0.16 & 0.05 & 0.46 & -0.08 & 0.0 & 0.0 & 0.06 & 0.0 & 0.05 & 0.01 & 0.0 & 0.01 & 0.0 & 0.05 & 0.0 & 0.01 & 0.0 & 0.0 & 0.01 \\
    \vspace{-0.49em}
    \raisebox{+.4\normalbaselineskip}[0pt][0pt]{fence} & 
    0.0 & 0.01 & 0.77 & 0.0 & -0.02 & 0.03 & 0.01 & 0.02 & 0.02 & 0.0 & 0.0 & 0.02 & 0.0 & 0.04 & 0.0 & 0.01 & 0.0 & 0.01 & 0.01 \\
    \vspace{-0.49em}
    \raisebox{+.4\normalbaselineskip}[0pt][0pt]{pole} & 
    0.0 & 0.01 & 0.64 & 0.0 & 0.02 & -0.07 & 0.04 & 0.02 & 0.05 & 0.01 & 0.0 & 0.02 & 0.0 & 0.04 & 0.0 & 0.01 & 0.01 & 0.02 & 0.01 \\
    \vspace{-0.49em}
    \raisebox{+.4\normalbaselineskip}[0pt][0pt]{t. light} & 
    0.0 & 0.0 & 0.55 & 0.0 & 0.03 & 0.1 & -0.11 & 0.04 & 0.06 & 0.01 & 0.0 & 0.01 & 0.0 & 0.01 & 0.0 & 0.0 & 0.0 & 0.05 & 0.0\\
    \vspace{-0.49em}
    \raisebox{+.4\normalbaselineskip}[0pt][0pt]{t. sign} & 
    0.0 & 0.0 & 0.67 & 0.0 & 0.03 & 0.07 & 0.03 & -0.04 & 0.02 & 0.0 & 0.0 & 0.01 & 0.0 & 0.06 & 0.0 & 0.01 & 0.0 & 0.02 & 0.01 \\
    \vspace{-0.49em}
    \raisebox{+.4\normalbaselineskip}[0pt][0pt]{veget.} & 
    0.0 & 0.0 & 0.11 & 0.0 & 0.0 & 0.01 & 0.26 & 0.0 & -0.3 & 0.06 & 0.0 & 0.0 & 0.0 & 0.02 & 0.0 & 0.0 & 0.0 & 0.17 & 0.0\\
    \vspace{-0.49em}
    \raisebox{+.4\normalbaselineskip}[0pt][0pt]{terrain} & 
    0.09 & 0.25 & 0.17 & 0.02 & 0.0 & 0.0 & 0.03 & 0.0 & 0.22 & -0.03 & 0.0 & 0.01 & 0.01 & 0.09 & 0.0 & 0.01 & 0.0 & 0.0 & 0.01 \\
    \vspace{-0.49em}
    \raisebox{+.4\normalbaselineskip}[0pt][0pt]{sky} & 
    0.0 & 0.0 & 0.08 & 0.0 & 0.01 & 0.07 & 0.0 & 0.01 & 0.03 & 0.0 & -0.58 & 0.0 & 0.0 & 0.0 & 0.01 & 0.0 & 0.05 & 0.11 & 0.0  \\
    \vspace{-0.49em}
    \raisebox{+.4\normalbaselineskip}[0pt][0pt]{person} & 
    0.0 & 0.03 & 0.49 & 0.0 & 0.0 & 0.0 & 0.02 & 0.0 & 0.01 & 0.0 & 0.0 & -0.04 & 0.0 & 0.25 & 0.0 & 0.03 & 0.0 & 0.02 & 0.04 \\
    \vspace{-0.49em}
    \raisebox{+.4\normalbaselineskip}[0pt][0pt]{rider} & 
    0.0 & 0.0 & 0.35 & 0.0 & 0.0 & 0.0 & 0.03 & 0.0 & 0.04 & 0.0 & 0.0 & 0.05 & -0.01 & 0.4 & 0.0 & 0.04 & 0.0 & 0.01 & 0.04\\
    \vspace{-0.49em}
    \raisebox{+.4\normalbaselineskip}[0pt][0pt]{car} & 
    0.01 & 0.0 & 0.1 & 0.0 & 0.0 & 0.0 & 0.0 & 0.0 & 0.0 & 0.0 & 0.0 & 0.02 & 0.0 & -0.81 & 0.0 & 0.02 & 0.0 & 0.0 & 0.03  \\
    \vspace{-0.49em}
    \raisebox{+.4\normalbaselineskip}[0pt][0pt]{truck} & 
    0.0 & 0.0 & 0.52 & 0.0 & 0.01 & 0.0 & 0.02 & 0.01 & 0.01 & 0.0 & 0.06 & 0.04 & 0.0 & 0.24 & -0.01 & 0.03 & 0.0 & 0.01 & 0.03\\
    \vspace{-0.49em}
    \raisebox{+.4\normalbaselineskip}[0pt][0pt]{bus} & 
    0.0 & 0.0 & 0.63 & 0.0 & 0.0 & 0.0 & 0.0 & 0.0 & 0.0 & 0.0 & 0.0 & 0.04 & 0.0 & 0.22 & 0.0 & -0.03 & 0.0 & 0.0 & 0.04 \\
    \vspace{-0.49em}
    \raisebox{+.4\normalbaselineskip}[0pt][0pt]{train} & 
    0.0 & 0.0 & 0.82 & 0.0 & 0.01 & 0.03 & 0.0 & 0.01 & 0.01 & 0.0 & 0.0 & 0.02 & 0.0 & 0.05 & 0.0 & 0.01 & -0.01 & 0.0 & 0.01 \\
    \vspace{-0.49em}
    \raisebox{+.4\normalbaselineskip}[0pt][0pt]{moto.} & 
    0.0 & 0.0 & 0.14 & 0.0 & 0.0 & 0.0 & 0.0 & 0.0 & 0.0 & 0.0 & 0.0 & 0.04 & 0.0 & 0.73 & 0.0 & 0.03 & 0.0 & -0.01 & 0.04 \\
    \vspace{-0.49em}
    \raisebox{+.4\normalbaselineskip}[0pt][0pt]{bicycle} & 
    0.01 & 0.02 & 0.28 & 0.0 & 0.0 & 0.0 & 0.02 & 0.0 & 0.02 & 0.0 & 0.0 & 0.04 & 0.01 & 0.45 & 0.0 & 0.03 & 0.0 & 0.01 & -0.04 \\

      \rotz{} &  
      \rotz{1} & \rotz{2} &  \rotz{3} & \rotz{4} &  \rotz{5} & \rotz{6} & \rotz{7} & \rotz{8} & \rotz{9} &  \rotz{10} & \rotz{11} & \rotz{12} & \rotz{13} & \rotz{14} & \rotz{15} & \rotz{16} & \rotz{17} & \rotz{18} & \rotz{19} 
      \end{tabular}
    }
}

 \caption{
 \textbf{Row-normalized confusion matrices}.
 }
 \label{fig:confusion_matrices_row_reordered1}
\end{figure*}

\subsection{Failure cases}
\label{sec:failure_cases}
\vspace{-1ex}

\begin{figure}[t!]
    \centering
        \footnotesize
         \centering
        \begin{tabular}{c@{}c@{}c@{}c@{}c}
            \raisebox{+1.1\normalbaselineskip}[0pt][0pt]{\rotatebox[origin=l]{90}{\small{GT \  \ \ \ \ \ Ours \ \ \ \ \ Input}}} &
            \includegraphics[height=4.5cm]{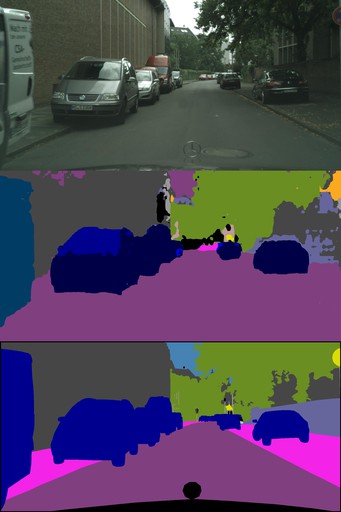} &
            \includegraphics[height=4.5cm]{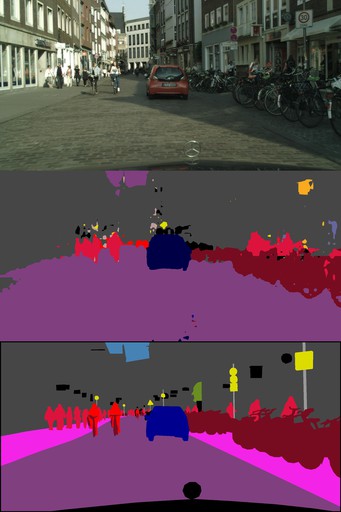} &
            \includegraphics[height=4.5cm]{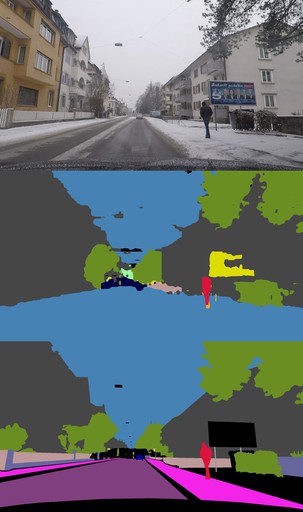} &
            \includegraphics[height=4.5cm]{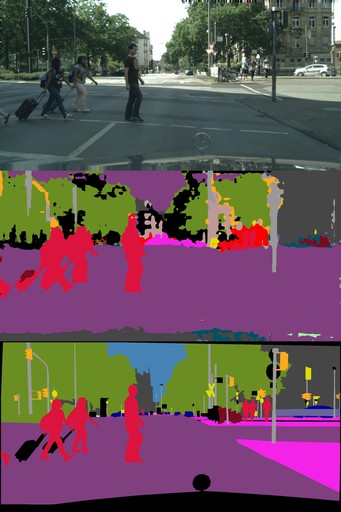} \\
            \multicolumn{5}{c}{\includegraphics[width=0.9\linewidth]{img/legend.jpg}} \\
            & {\small (a) mixed labels} & {\small(b) good shape,} & \multicolumn{2}{c}{{\small (c) sky $\leftrightarrow$ road confusion}} \\
            & & {\small wrong class} &  \\
         \end{tabular}
    \caption{\textbf{Failure cases}. \textbf{(a)} Due to the noise in the training data (discussed in Section~\ref{sec:failure_cases}; images and LiDAR point clouds come from the Waymo Open~\cite{sun2020scalability} dataset), \ours sometimes predicts multiple pseudo-labels inside the same object (here different shades of blue inside the car on the left). 
    \textbf{(b)}
    Objects that belong to the same semantic category (\eg, \emph{cars}) might end-up clustered into different pseudo-classes due to differences in appearance (\eg, a separate pseudo-class that corresponds to the rear of the cars).
    \textbf{(c)} The \emph{road}$\leftrightarrow$\emph{sky} misplacement/confusion is caused by the absence of sky-occupied labeled pixels at training as they are not covered by the LiDAR data. Therefore, the model assigns the most common label to the sky, which is the pseudo-label that corresponds to the road. This leads to either predicting the road as \textit{sky} (third column), or predicting sky as \textit{road} (fourth column), depending on the outcome of the Hungarian matching.
    }
    \label{fig:failure}
\end{figure}

The main limitations of our Drive\&Segment approach are discussed in Section~4.4 of the main paper. Here, we show some qualitative examples of these failure modes and discuss more thoroughly their roots.

The first limitation of \ours is the complete absence of pseudo-labeled training data for the \emph{sky} class.
This is because 
the LiDAR data do not capture the sky.
As a consequence, our models learn to classify the \emph{sky} pixels as \emph{road} (see the ``sky" row of the confusion matrix in Figure~\ref{fig:confusion_segmenter_row_reordered}),
which is the most dominant (pseudo-)label in the data.
We provide examples of this behavior in  Figure~\ref{fig:failure}(c).

The second most common failure mode is inherited from the object proposal method that relies only on geometry-derived features. 
Specifically, the segment proposal method might over-segment an object, 
potentially causing different object parts being assigned to different pseudo-labels.
The class majority voting in our refinement stage does not always rectify this issue.
As a consequence, our models might learn to make predictions that mix multiple pseudo-labels in one object. 
For example, see Figure~\ref{fig:failure}(a) where the car is mixed with the \emph{truck} class.

Finally, our LiDAR-based proposal method groups all points from the ground plane into a single segment, without being able to distinguish the various ground-plane classes (\eg, \textit{road}, \textit{sidewalk} and \textit{terrain}) that are defined in the image domain. Figure~\ref{fig:failure} provides examples of this failure mode. This phenomenon is also well visible in Figure~\ref{fig:confusion_segmenter_row_reordered}.

\section{Additional qualitative results}

\subsection{Qualitative comparison to previous work}
\label{sec:qual}
\begin{figure*}[t]
    \centering
    \includegraphics[width=1.0\linewidth]{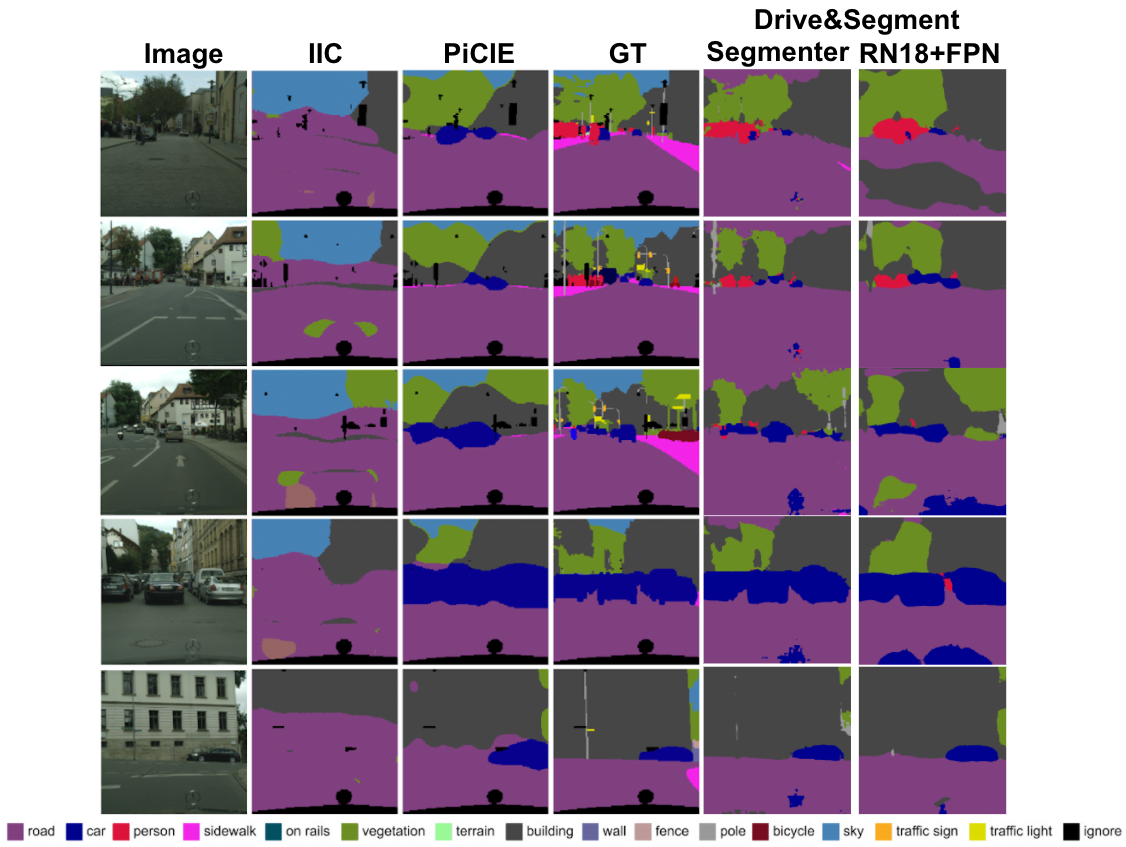}
    \vspace{-4ex}
    \caption{\textbf{Qualitative comparison of PiCIE~\cite{cho2021picie}, IIC~\cite{ji2019invariant} and our \ours approach on PiCIE \emph{training} samples.} For a fair comparison, we use the same visualization procedure as in~\cite{cho2021picie}. 
    Results are shown on center-cropped Cityscapes training images. Note that our method is able to capture objects' contours much better and to segment categories such as \emph{person} that are not visible in IIC or PiCIE results.}
    \label{fig:ours_vs_picie_crops}
\end{figure*}

We show a qualitative comparison with IIC~\cite{ji2019invariant} and PiCIE~\cite{cho2021picie} in Figure~\ref{fig:ours_vs_picie_crops}. 
For a fair comparison, we 
use the same samples and the visualization protocol as in \cite{cho2021picie}. 
Note that these samples come from the PiCIE and IIC training set, namely from the \emph{train} set of the Cityscapes~\cite{Cordts2016Cityscapes} dataset, while for our method these are only test samples. %
In Figure~\ref{fig:ours_vs_picie_crops}, note how our \ours %
is able to segment the \emph{person} class, while neither IIC nor PiCIE are capable to do so.

\subsection{Qualitative Results}
\label{sec:qual_ours}
\begin{figure*}[t!]
    \begin{minipage}[t]{\columnwidth}
        \footnotesize
         \centering
         \begin{tabular}{c@{}c@{}c}
         Input & Ground Truth & Drive\&Segment (Ours) \\
         \vspace{-3pt}
         \includegraphics[width=0.33\textwidth]{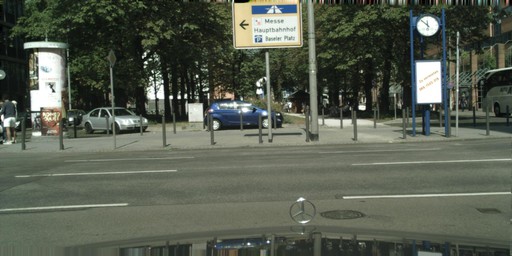} &
         \includegraphics[width=0.33\textwidth]{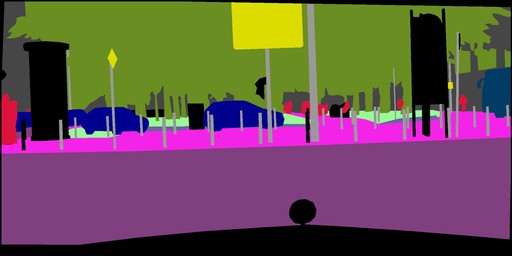} &
         \includegraphics[width=0.33\textwidth]{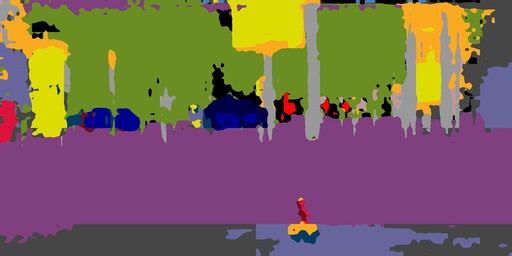} \\
         \vspace{-3pt}
         \includegraphics[width=0.33\textwidth]{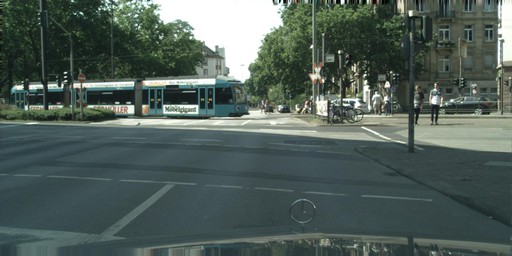} &
         \includegraphics[width=0.33\textwidth]{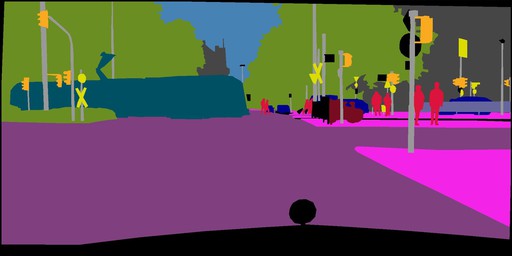} &
         \includegraphics[width=0.33\textwidth]{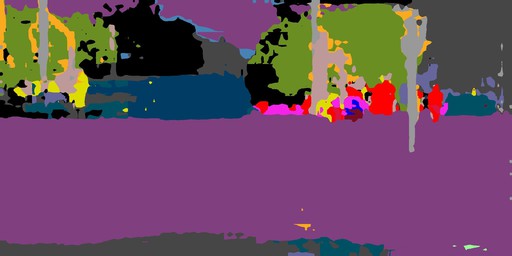} \\
         \vspace{-3pt}
         \includegraphics[width=0.33\textwidth]{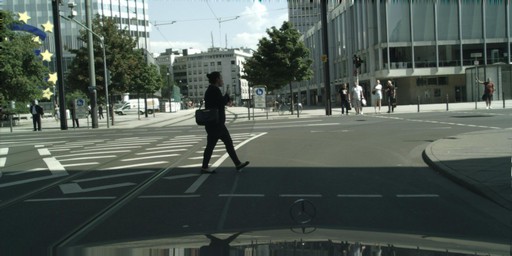} &
         \includegraphics[width=0.33\textwidth]{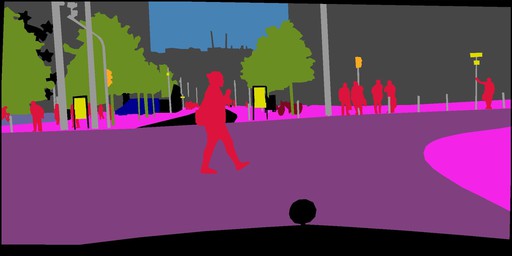} &
         \includegraphics[width=0.33\textwidth]{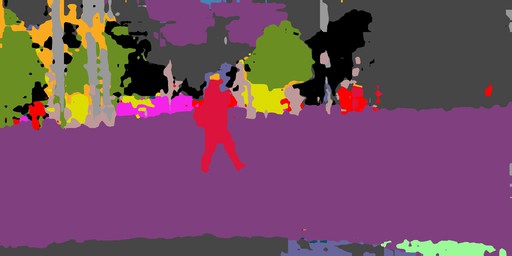} \\
         \vspace{-3pt}
         \includegraphics[width=0.33\textwidth]{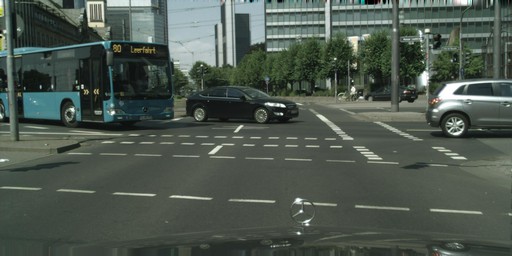} &
         \includegraphics[width=0.33\textwidth]{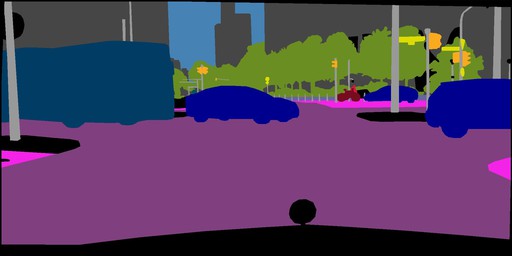} &
         \includegraphics[width=0.33\textwidth]{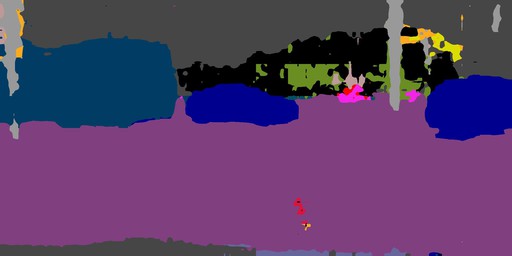} \\
         \vspace{-3pt}
         \includegraphics[width=0.33\textwidth]{img/supplementary/selection/frankfurt_000001_055062_leftImg8bit_orig.jpg} &
         \includegraphics[width=0.33\textwidth]{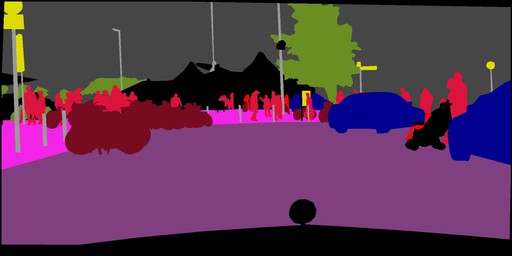} &
         \includegraphics[width=0.33\textwidth]{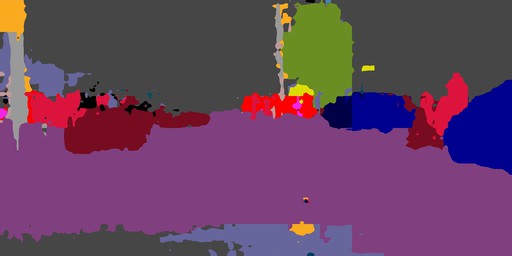} \\
         \vspace{-3pt}
         \includegraphics[width=0.33\textwidth]{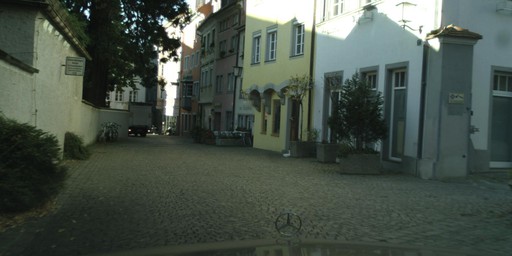} &
         \includegraphics[width=0.33\textwidth]{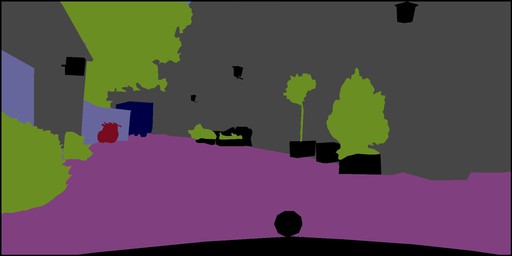} &
         \includegraphics[width=0.33\textwidth]{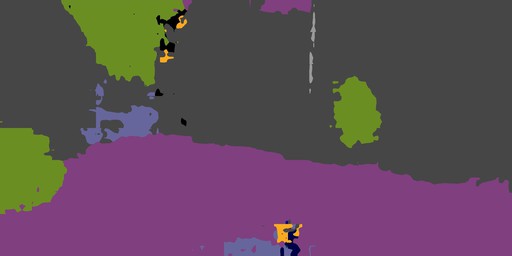} \\
         \vspace{-3pt}
         \includegraphics[width=0.33\textwidth]{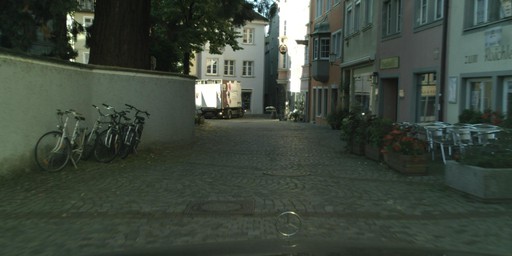} &
         \includegraphics[width=0.33\textwidth]{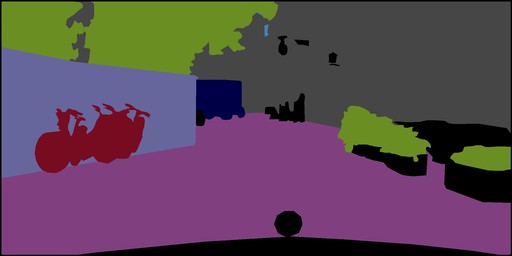} &
         \includegraphics[width=0.33\textwidth]{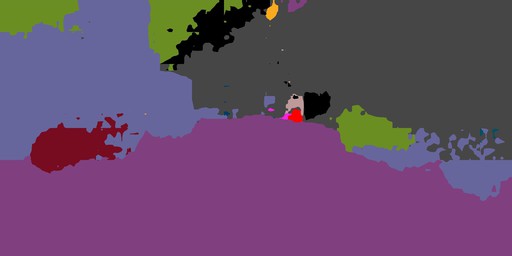} \\
         \vspace{-3pt}
         \includegraphics[width=0.33\textwidth]{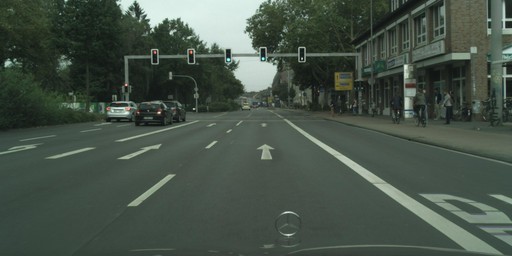} &
         \includegraphics[width=0.33\textwidth]{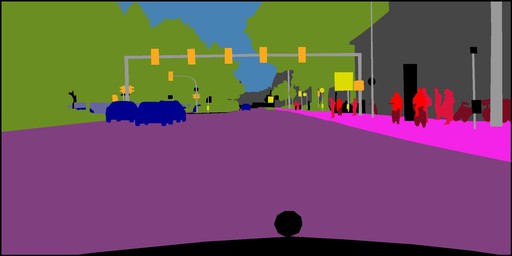} &
         \includegraphics[width=0.33\textwidth]{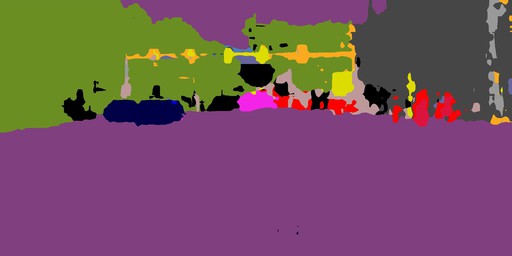} \\
         \end{tabular}
    \end{minipage}\\
    \includegraphics[width=0.94\linewidth]{img/legend.jpg}
    \vspace*{-2ex}
    \caption{
    \textbf{Qualitative results for unsupervised semantic segmentation using our \ours approach on the validation split of the Cityscapes dataset.}
    The matching
between our pseudo-classes and the set of ground-truth classes is obtained using the Hungarian algorithm.
    }
    \label{fig:qualitative_supp}
    \vspace{-1ex}
\end{figure*}

\begin{figure*}[t!]
    \begin{minipage}[t]{\columnwidth}
        \footnotesize
         \centering
         \begin{tabular}{c@{}c@{}c}
         Input & Ground Truth & Drive\&Segment (Ours) \\
         \vspace{-3pt}
         \includegraphics[width=0.33\textwidth]{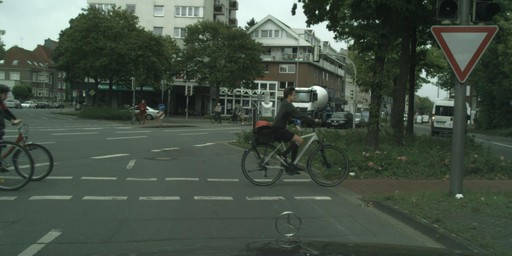} &
         \includegraphics[width=0.33\textwidth]{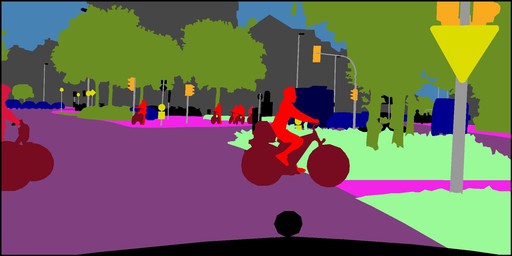} &
         \includegraphics[width=0.33\textwidth]{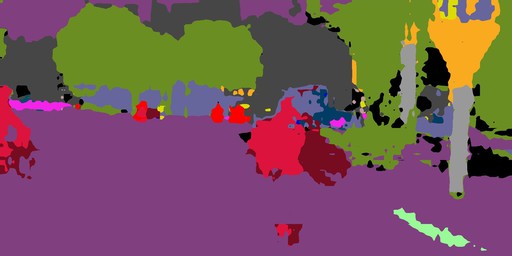} \\
         \vspace{-3pt}
         \includegraphics[width=0.33\textwidth]{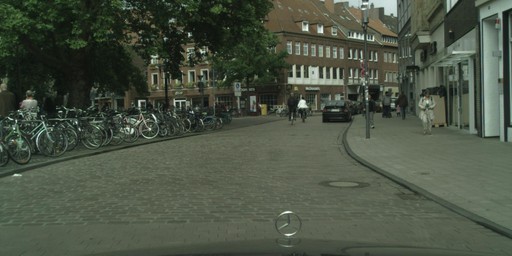} &
         \includegraphics[width=0.33\textwidth]{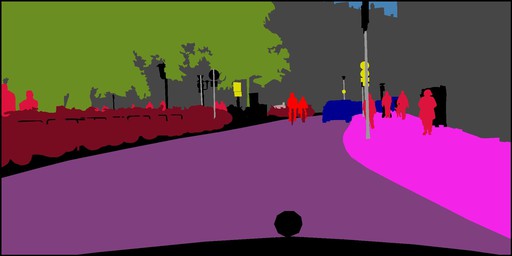} &
         \includegraphics[width=0.33\textwidth]{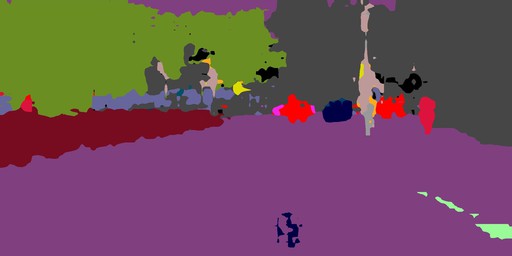} \\
         \vspace{-3pt}
         \includegraphics[width=0.33\textwidth]{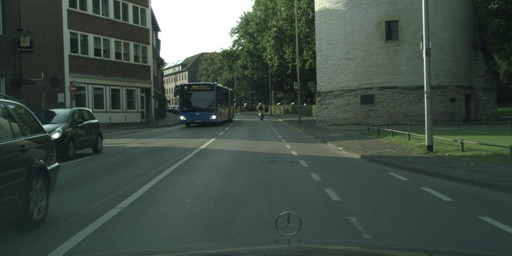} &
         \includegraphics[width=0.33\textwidth]{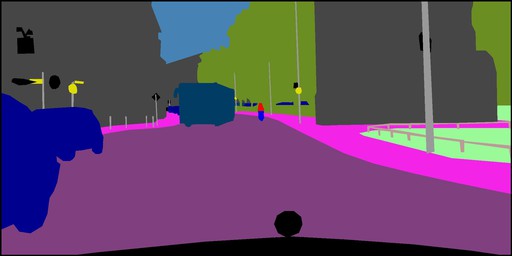} &
         \includegraphics[width=0.33\textwidth]{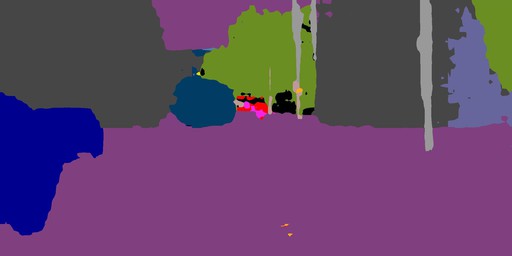} \\
         \vspace{-3pt}
         \includegraphics[width=0.33\textwidth]{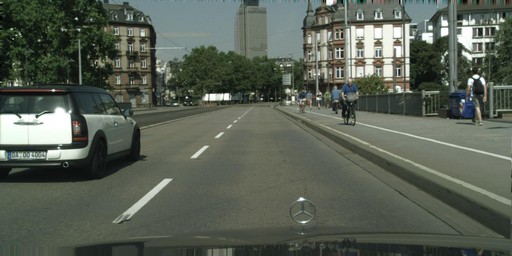} &
         \includegraphics[width=0.33\textwidth]{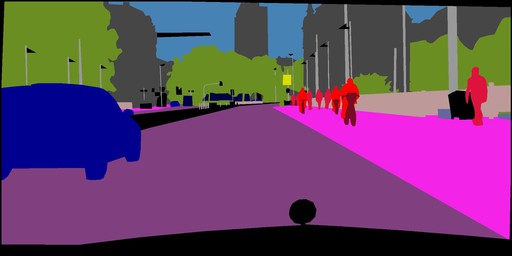} &
         \includegraphics[width=0.33\textwidth]{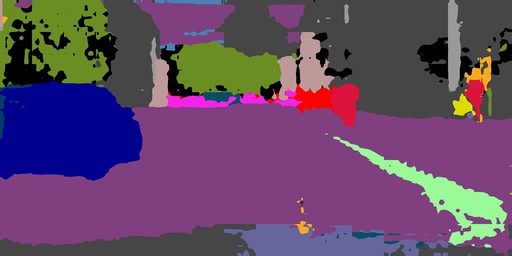} \\
         \vspace{-3pt}
         \includegraphics[width=0.33\textwidth]{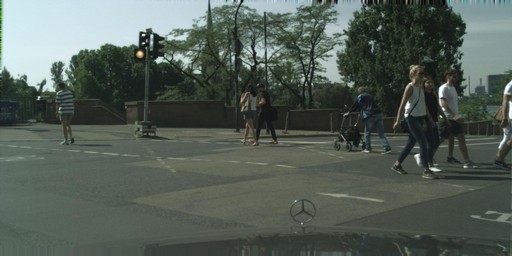} &
         \includegraphics[width=0.33\textwidth]{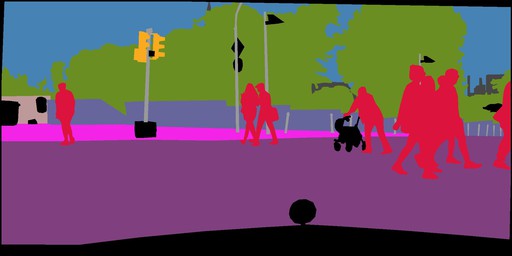} &
         \includegraphics[width=0.33\textwidth]{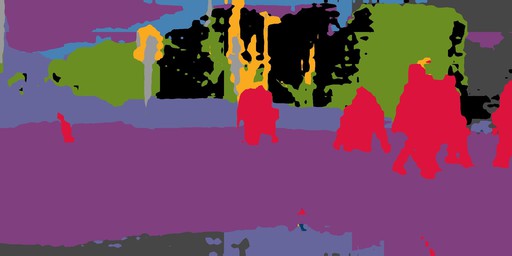} \\
         \vspace{-3pt}
         \includegraphics[width=0.33\textwidth]{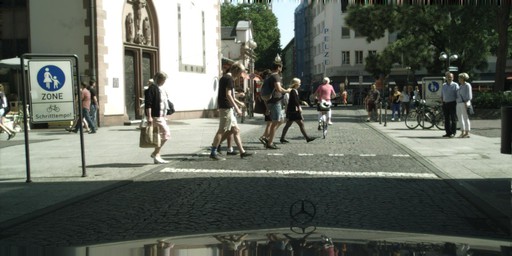} &
         \includegraphics[width=0.33\textwidth]{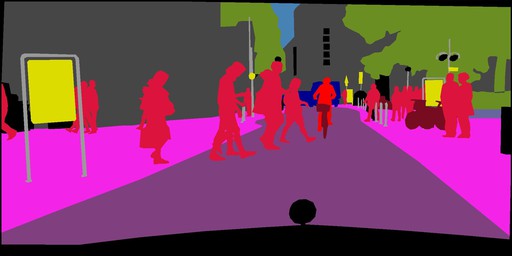} &
         \includegraphics[width=0.33\textwidth]{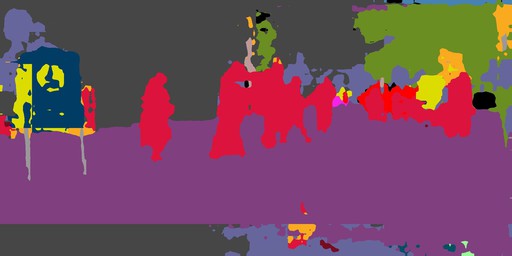} \\
         \vspace{-3pt}
         \includegraphics[width=0.33\textwidth]{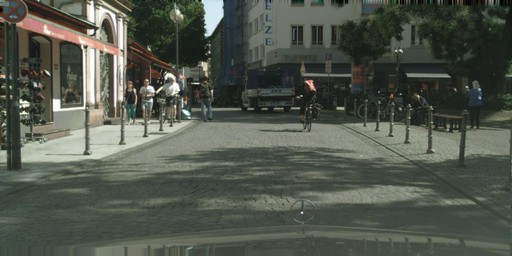} &
         \includegraphics[width=0.33\textwidth]{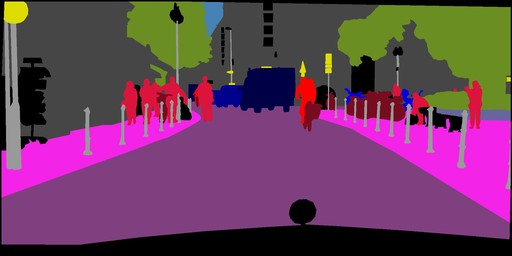} &
         \includegraphics[width=0.33\textwidth]{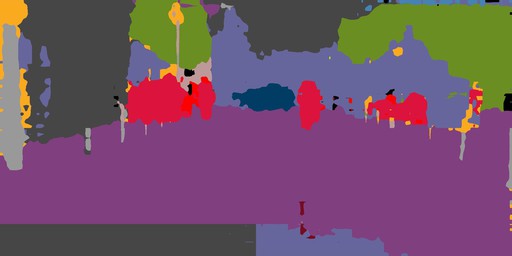} \\
         \vspace{-3pt}
         \includegraphics[width=0.33\textwidth]{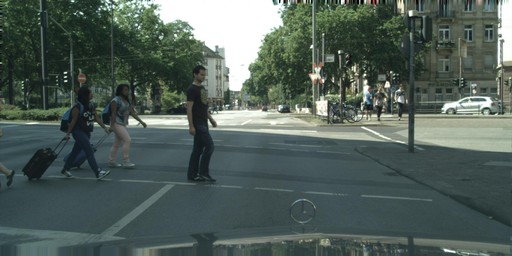} &
         \includegraphics[width=0.33\textwidth]{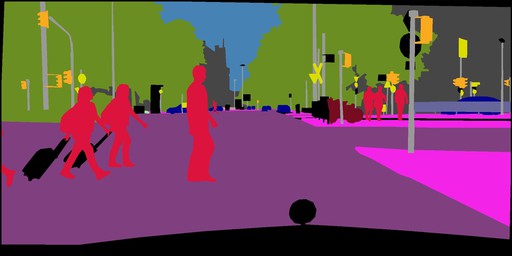} &
         \includegraphics[width=0.33\textwidth]{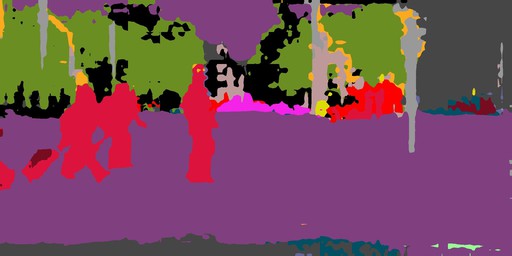} \\
         \end{tabular}
    \end{minipage}\\
    \includegraphics[width=0.94\linewidth]{img/legend.jpg}
    \vspace*{-2ex}
    \caption{
    \textbf{Qualitative results for unsupervised semantic segmentation using our \ours approach on the validation split of the Cityscapes dataset.} The matching
between our pseudo-classes and the set of ground-truth classes is obtained using the Hungarian algorithm.
    }
    \label{fig:qualitative_supp2}
    \vspace{-1ex}
\end{figure*}

\begin{figure*}[t!]
    \includegraphics[width=1.0\textwidth]{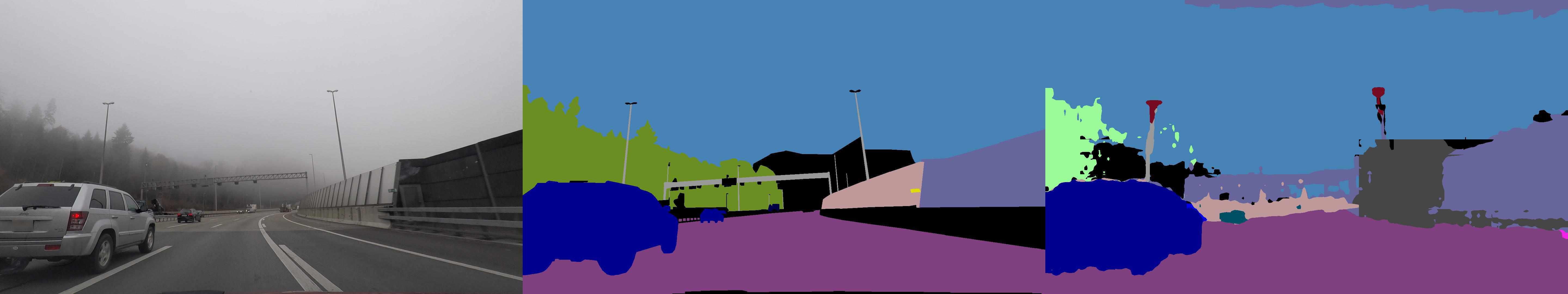} \\
    \includegraphics[width=1.0\textwidth]{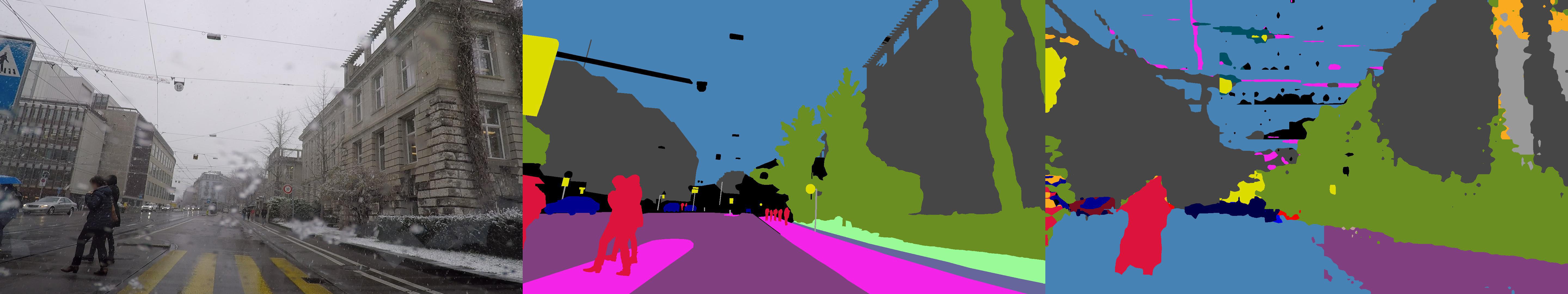} \\
    \includegraphics[width=1.0\textwidth]{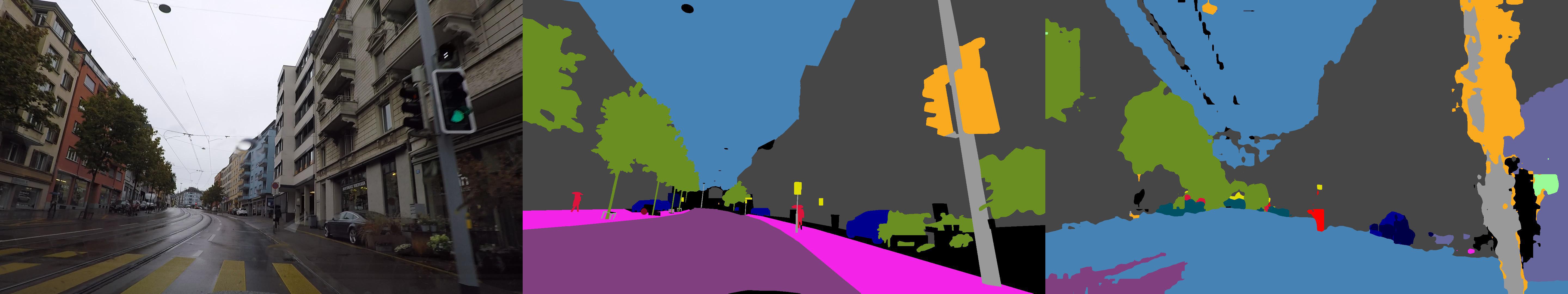} \\
    \includegraphics[width=1.0\textwidth]{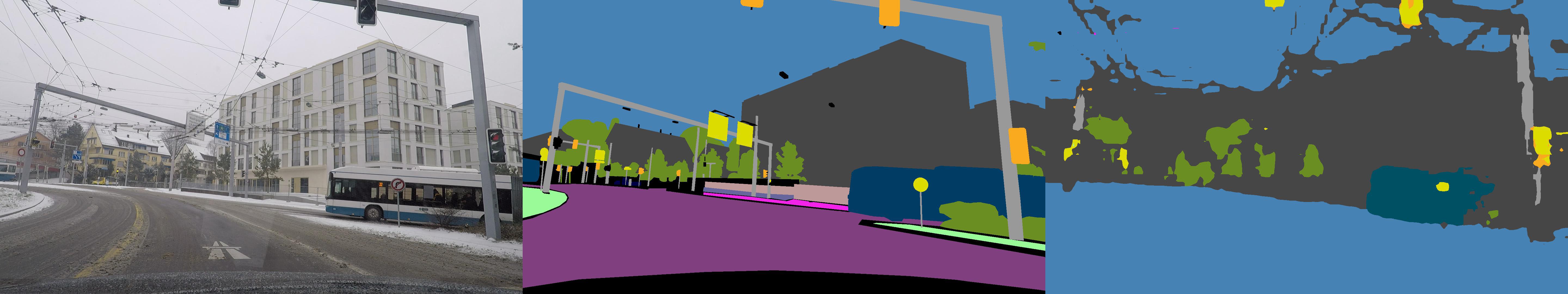} \\
    \includegraphics[width=1.0\textwidth]{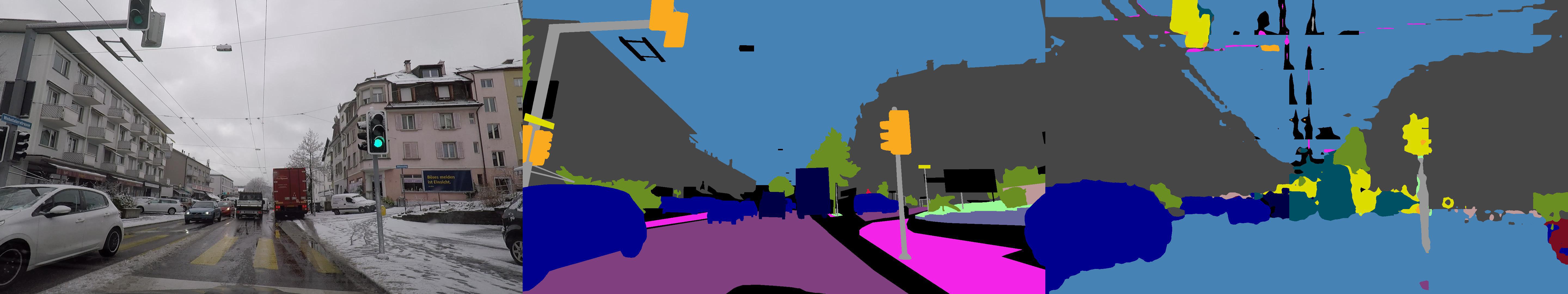} \\
    \includegraphics[width=1.0\textwidth]{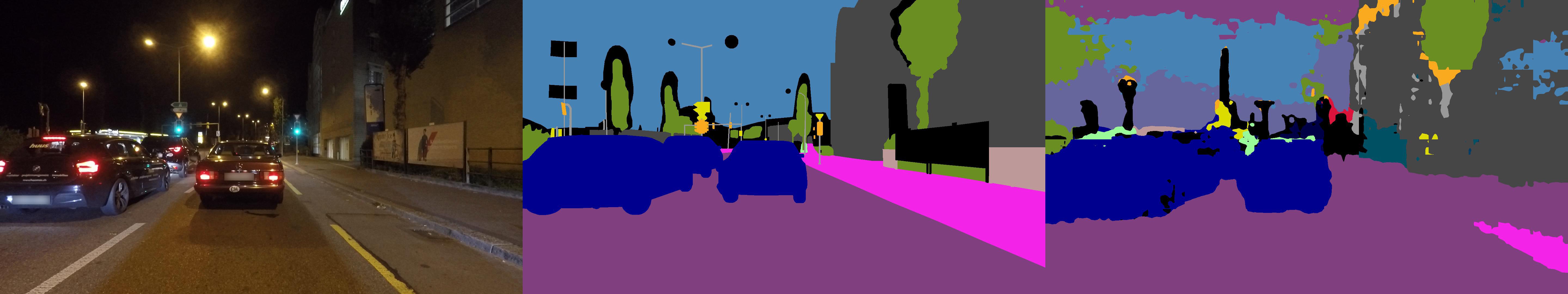} \\
    \includegraphics[width=1.0\textwidth]{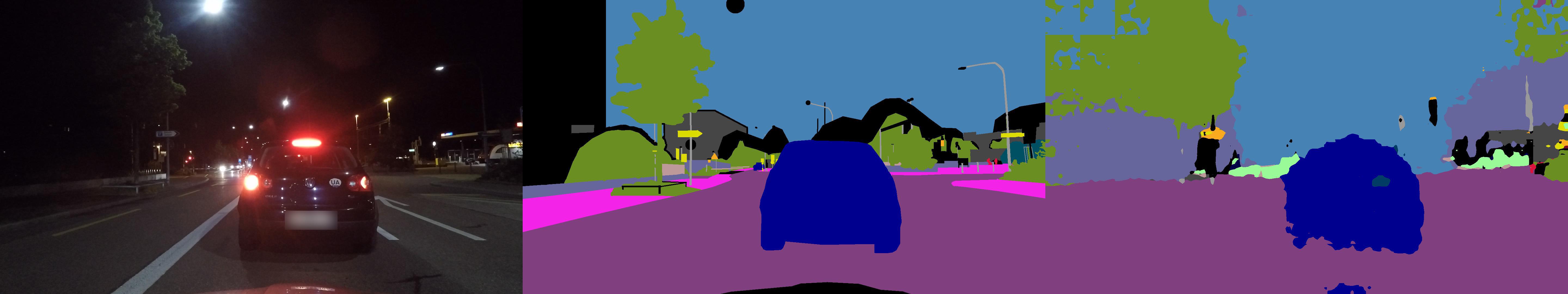} \\
    \includegraphics[width=\linewidth]{img/legend.jpg} \\
    {\small \hspace*{1.7cm} Input \hspace{2.2cm} Ground Truth \hspace{1.5cm} \ours (Ours)}
\vspace*{-8pt}
\caption{\textbf{Waymo Open Dataset \emph{day} $\rightarrow$ ACDC~\cite{SDV21} \{\emph{fog, rain, snow, night}\}.} Qualitative results of our \ours model trained on the daytime images from the Waymo Open Dataset and used to segment samples from the ACDC~\cite{SDV21} dataset with various adverse conditions. 
In rows 2-5 the \emph{ground} is incorrectly segmented as \emph{sky}. This failure mode is further discussed in Section~\ref{sec:failure_cases}. %
}
\label{fig:qualitative_acdc}
\vspace{-1ex}
\end{figure*}

In Figures~\ref{fig:qualitative_supp} and~\ref{fig:qualitative_supp2},
we report \ours predictions on Cityscapes validation images. In spite of the domain gap between the training dataset (Waymo Open Dataset~\cite{sun2020scalability} with images from US cities) and the Cityscapes test set, our approach produces convincing results. 
Furthermore, in Figure~\ref{fig:qualitative_acdc}, we report qualitative results of our method pretrained on \emph{daytime-only} images and evaluated on \emph{out-of-training-distribution} splits of ACDC~\cite{SDV21}, \eg, \emph{night}, \emph{snow} or \emph{fog}.
We discuss the main failure modes in  Section~\ref{sec:failure_cases}.

\begin{table}[t!]
\caption{\textbf{Supervised fine-tuning on the Cityscapes~\cite{Cordts2016Cityscapes} semantic segmentation task.}
Results report mean Intersetion over Union (mIoU).
We fine-tune the pre-trained ResNet18+FPN networks on either the entire Cityscapes training split (`Full Cityscapes') or only 100 images from the training split (`Low-shot')~\cite{french2020semi} and test on the Cityscapes validation split.
`Linear' fine-tunes only the last linear layer, `Decoder+Linear' fine-tunes the FPN decoder and the last linear layer, and `End-to-End' fine-tunes the entire network.
}
\centering
{
\setlength{\extrarowheight}{1.5pt}%
{
\begin{tabular}{l <{\hspace{0.5em}} | >{\hspace{0.5em}} c <{\hspace{0.5em}} | >{\hspace{0.5em}} c <{\hspace{0.5em}} | >{\hspace{0.5em}} c <{\hspace{0.5em}}}
\toprule
& \multicolumn{2}{c|}{Full Cityscapes} & \multicolumn{1}{c}{Low-shot}\\
Pre-training & Linear & Decoder+Linear & End-to-End\\
\hline
\hspace{1mm}PiCIE~\cite{cho2021picie}  & 17.4 & 29.5 & 30.4 \\
\hspace{1mm}Imagenet (supervised)                 & 25.7 & 41.9 & 48.2 \\
\hspace{1mm}\ours                      & \textbf{36.4} & \textbf{46.4} & \textbf{49.2} \\ 
\bottomrule
\end{tabular}
}
}
\vspace{-0ex}
\label{tab:finetune_CS_v2}
\end{table}

\section{Evaluating \ours with supervised fine-tuning} 
\label{sec:fine}

The goal of our work is to train image segmentation models without any human annotation.
Here, we evaluate with some preliminary experiments the applicability of the proposed \ours method on a different but related task, that of \emph{self-supervised pre-training} of semantic segmentation networks (i.e., self-supervised feature learning).
Specifically, we take the ResNet18\Plus FPN model trained with \ours, 
replace its last linear prediction layer with a new layer that has as many outputs as classes in Cityscapes ($19$), and
fine-tune the resulting network on the Cityscapes~\cite{Cordts2016Cityscapes} dataset using available human annotations. We compare against (a) using PiCIE~\cite{cho2021picie} for self-supervised pre-training and (b) \emph{supervised} pre-training on ImageNet~\cite{deng2009imagenet}.

We evaluate the different pre-training approaches with three 
fine-tuning setups.
The first setup is to freeze both the ResNet18 backbone and the FPN decoder (i.e., keep their pre-trained weights fixed) and fine-tune only the last linear prediction layer.
The second setup is to freeze only the ResNet18 backbone and fine-tune both the FPN decoder and the last linear layer.
In both cases, we train on the entire training split ($2975$ images) of Cityscapes.
The goal of these first two setups is to evaluate the quality of the pre-trained ResNet18\Plus FPN (1st setup) or ResNet18 (2nd setup) features as they are.
The third setup targets the \emph{low-shot scenario}: the segmentation network is  fine-tuned end-to-end using only $100$ Cityscapes training images (we consider three random splits of 100 images from~\cite{french2020semi}).
The purpose of this setup is to evaluate the strength of the pre-trained network in a regime where only a few annotations are available for fine-tuning. 

In the first two setups we train for 
$40k$ iterations, and we train for $4k$ in the low-shot setup.
In all setups, we use SGD with momentum set to $0.9$, weight decay to $0.0005$ and mini-batches of size $8$.
During training we use random image scaling (by a ratio in $[0.5, 2.0]$), random cropping (with size $769$) and horizontal flipping. At test time, we use the original image size and horizontal flip augmentations.
The learning rates were tuned for each fine-tuning setup and each evaluated method separately. 

We report results in Table~\ref{tab:finetune_CS_v2}.
Although our method was not designed or optimized for self-supervised feature pre-training, it still provides promising results that surpass both PiCIE and ImageNet pre-training.

\end{document}